\documentclass[10pt,final,journal,twocolumn]{IEEEtran}
\usepackage[caption=false,font=footnotesize]{subfig}
\usepackage[boxed,linesnumbered,algoruled,noline]{algorithm2e}
\usepackage{setspace}
\usepackage{graphicx}
\usepackage{amsmath,amsthm,amssymb}
\usepackage{wasysym}
\usepackage{bbm,dsfont}
\usepackage{hyperref}
\usepackage{multirow}
\usepackage{booktabs}
\usepackage{url}
\usepackage{color}

\graphicspath{{./}}

\newcommand{\etal}{et al.\thinspace}

\newcommand{\eq}[1]{Eq.\thinspace\eqref{#1}}
\newcommand{\fig}[1]{Fig.\thinspace\ref{#1}}

\newcommand{\floor}[1]{\lfloor#1\rfloor}

\newcommand{\card}[1]{\left|#1\right|}
\DeclareMathOperator*{\argmax}{arg\,max}

\newcommand{\neuron}{\operatorname{neuron}}
\newcommand{\sos}{\textsc{sum-of-sum}}
\newcommand{\som}{\textsc{sum-of-max}}

\title{Combating Corrupt Messages in Sparse Clustered Associative Memories}
\author{
    \IEEEauthorblockN{Zhe~Yao\IEEEauthorrefmark{1}, Vincent~Gripon\IEEEauthorrefmark{2} and Michael~G.~Rabbat\IEEEauthorrefmark{1}}
    \thanks{\IEEEauthorrefmark{1}~Z.~Yao and M.G.~Rabbat are with the Department of Electrical and Computer Engineering, McGill University, Montr\'{e}al, QC, Canada. Email: \href{mailto:zhe.yao@mail.mcgill.ca}{zhe.yao@mail.mcgill.ca}, \href{mailto:michael.rabbat@mcgill.ca}{michael.rabbat@mcgill.ca}.
    }
    \thanks{\IEEEauthorrefmark{2}~V.~Gripon is with the Electronics Department, T\'{e}l\'{e}com Bretagne, Brest, France. Email: \href{mailto:vincent.gripon@telecom-bretagne.eu}{vincent.gripon@telecom-bretagne.eu}.
    }
}

\begin{document}
\maketitle
\begin{abstract}
In this paper we analyze and extend the neural network based associative memory proposed by Gripon and Berrou.
This associative memory resembles the celebrated Willshaw model with an added partite cluster structure.
In the literature, two retrieving schemes have been proposed for the network dynamics, namely \sos{} and \som{}.
They both offer considerably better performance than Willshaw and Hopfield networks, when comparable retrieval scenarios are considered.
Former discussions and experiments concentrate on the erasure scenario, where a partial message is used as a probe to the network, in the hope of retrieving the full message.
In this regard, \som{} outperforms \sos{} in terms of retrieval rate by a large margin.
However, we observe that when noise and errors are present and the network is queried by a corrupt probe, \som{} faces a severe limitation as its stringent activation rule prevents a neuron from reviving back into play once deactivated.
In this manuscript, we categorize and analyze different error scenarios so that both the erasure and the corrupt scenarios can be treated consistently.
We make an amendment to the network structure to improve the retrieval rate, at the cost of an extra scalar per neuron.
Afterwards, five different approaches are proposed to deal with corrupt probes.
As a result, we extend the network capability, and also increase the robustness of the retrieving procedure.
We then experimentally compare all these proposals and discuss pros and cons of each approach under different types of errors.
Simulation results show that if carefully designed, the network is able to preserve both a high retrieval rate and a low running time simultaneously, even when queried by a corrupt probe.
\end{abstract}

\section{Introduction\label{sec:introduction}}
\subsection{Background}
Associative memories are devices that map pairs of input-output patterns and retrieve information from its context directly. 
Thus they behave differently than traditional memory systems where explicit addresses are required to retrieve the content,
There are two phases in using an associative memory: storing and retrieving.
In the storing phase, the task is to store (learn) all messages of interest into the memory.
In the retrieving phase, given a probe, a modified version of a particular message, one is expected to retrieve (decode) the originally stored version reliably and efficiently.

Associative memories have a variety of uses in different fields, e.g., communication networks~\cite{kaxiras2005ipstash}, signal and image processing~\cite{valle2009class}, database engines~\cite{lin1976rares}, anomaly detection systems~\cite{bu2004camnids}, compression algorithms~\cite{lin2000camlz} and face recognition systems~\cite{zhang2005gabor}, to name a few.
Some implementations provide query time independent of the number of stored messages.

Neural networks are among the most popular approaches to implement associative memories, e.g., linear associators~\cite{anderson1988neurocomputing,anderson1993neurocomputing2}, Willshaw networks~\cite{willshaw1969non,willshaw1971models} are early examples of such attempts.
In early 1980's, the seminal work of Hopfield~\cite{hopfield1982neural,hopfield1984neurons} on associative memories brought back research interest for the neural network community.
For the history and developments of associative memories, see Palm's recent survey~\cite{palm2013neural} and the references therein.

Quite recently, Gripon and Berrou propose a new family of sparse neural networks for associative memories~\cite{gripon2011simple,gripon2011sparse} which we refer to as the \emph{Clustered Sparse Associative Memory} (\textbf{CSAM}).
In short, an CSAM is a modification of Willshaw networks with partite cluster structures.
It resembles the model proposed by Moopenn~\etal~\cite{moopenn1987electronic} with the original retrieving scheme, \sos{}~\cite{gripon2011sparse}.
But it also allows for neuron self excitations, as well as a new retrieving scheme \som{}~\cite{gripon2012nearly} which improves the retrieval rate by a large margin.
A brief description of CSAMs and these two basic retrieving schemes will be given in Section~\ref{sec:gbnn}.

\subsection{Related Work}
Gripon and Berrou propose the network structure in~\cite{gripon2011simple}.
In~\cite{gripon2011sparse}, they show that using the same amount of storage, CSAMs outperform Hopfield networks in diversity (the number of patterns a network can store for a targeted performance), capacity (the maximum amount of stored information in bits for a targeted performance) and efficiency (the ratio between capacity and the amount of information in bits consumed by the network when capacity reaches its maximum) simultaneously.
They later interpret CSAMs using the formalism of error correcting codes~\cite{gripon2012nearly} and propose a new decoding scheme called \som{}, which significantly decreases retrieval error.
Jiang~\etal~\cite{jiang2012learning} modify CSAMs to store long sequences by incorporating directed links.
Aboudib~\etal~\cite{aboudib2013study} extend the structure so that messages of different lengths can be stored in the same network.
They also summarize criteria to build possible retrieving schemes and study the number of iterations required by each scheme.
Yao~\etal~\cite{yao2013bogus} discover a previously overlooked problem that the network may converge to a bogus fixed point and propose heuristics to mitigate the issue.
A novel post-processing algorithms is also developed, customized to the partite structure of CSAMs, which brings notably better retrieval rates than the standard \som{} scheme.

Aside from the architectural and algorithmic aspects of the network mentioned above, efficient implementations and applications are being developed as well.
Jarollahi~\etal~\cite{jarollahi2012architecture} use the \emph{field programmable gateway array} (FPGA) to implement \sos{} on a small sized network.
Later in~\cite{jarollahi2013reduced}, they implement \som{} which runs $1.9\times$ faster, thanks to bitwise operations replacing the resource demanding summation and comparison units required by \sos{}.
The same group of authors also develop a content addressable memory in~\cite{jarollahi2013lowpower} saving $90\%$ of the energy consumption.
Larras~\etal~\cite{larras2013analog} develop an analog version of the network, which consumes $1165\times$ less energy.
Meanwhile, it is $2\times$ more efficient in terms of both speed and circuit surface, comparing with an equivalent digital circuit.
After analyzing the convergence and computational properties of both \sos{} and \som{}, Yao~\etal~\cite{yao2013gpugbnn} propose a hybrid scheme and successfully implement the network on a GPU.
An acceleration of $900\times$ is witnessed without any loss of accuracy.

\subsection{Contributions}
Former discussions and experiments with CSAMs concentrate on the erasure scenario, where a partial message is used as a probe to the network, in the hope of retrieving the full message.
In this regard, \som{} outperforms \sos{} in terms of retrieval rate by a large margin~\cite{yao2013gpugbnn}.
However, we argue that real world data applications may include more challenging scenarios, where inputs contain errors. 
\som{} faces a severe limitation in such scenarios as its stringent activation rule prevents a neuron from being reactivated.

The contributions of this manuscript are:
\begin{enumerate}
	\item We categorize and analyze different errors when using an CSAM so that the erasure scenario and the corruption scenario can be treated consistently.
	\item We make an amendment to the existing network structure to improve the retrieval rate further at the cost of an extra scalar per neuron.
	\item We propose five different approaches to deal with errors, thus extend the network capability and also increase the robustness of retrieving procedure.
\end{enumerate} 

The numerical experiments in Section~\ref{sec:experiment} compare all proposals and modifications under different types of errors to see the pros and cons of each approach.
Both simulated data and the famous USPS dataset are tested.
Experimental results show that if carefully designed, the network is able to preserve both a high retrieval rate and a low running time simultaneously, even when queried by a corrupt probe.

\subsection{Paper Organization}
The rest of the paper is organized as follows.
Section~\ref{sec:gbnn} reviews the structure of CSAMs and two existing retrieving algorithms, \sos{} and \som{}.
We explain the reason in brief why \som{} should be favored over \sos{}.
Section~\ref{sec:error} categorizes and analyzes different errors.
We discuss three basic types of errors and how \sos{} and \som{} behave against them.
Section~\ref{sec:amendment} describes the structural amendment we make to CSAMs so that retrieving bias can be mitigated and better retrieval rates can be achieved, at the cost of an extra scalar for each neuron.
In Section~\ref{sec:proposal}, five different approaches are proposed along with their pseudo code implementations.
Section~\ref{sec:experiment} compares numerically all proposals under different errors separately in details using both simulated and real world data.
The manuscript concludes in Section~\ref{sec:summary}.

\section{Sparse Clustered Associative Memories\label{sec:gbnn}}
\subsection{Structure}
The structure of an CSAM~\cite{gripon2011simple} is closely coupled with stored patterns.
Consider a $C$-symbol tuple, $(m_1, m_2, \cdots, m_C)$, with each symbol $m_c$ taking $L$ possible values, i.e., $m_c=x_l, l=1,2,\cdots,L$.
We call such a tuple a \emph{message}.
In this setup, a network of $n=CL$ neurons is used, with $C$ clusters corresponding to different symbols, each having $L$ neurons representing $x_l$.
If $m_c=x_l$, the $l$\textsuperscript{th} neuron in the $c$\textsuperscript{th} cluster, $\neuron(c,l)$, activates.
Since a symbol can only take one value at a time, for a given message, in each cluster, only one single neuron activates accordingly.
The locations of the active neurons express a particular message.

\begin{figure}
\centering
\includegraphics[scale=.35]{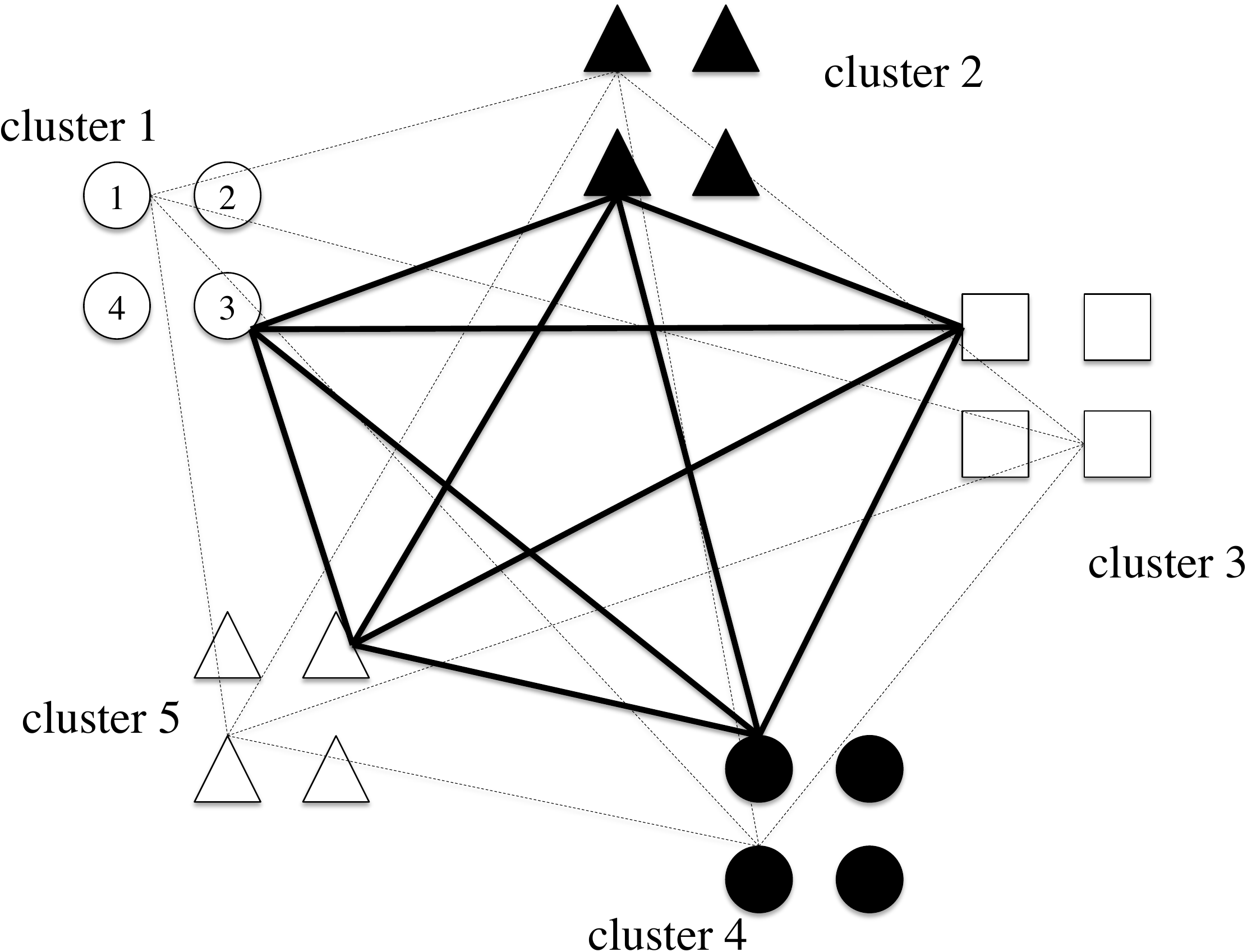}
\caption{
	A toy network with $5$ clusters of $4$ neurons each.
	We number clusters and neurons clockwise.
	Two messages are stored.
}
\label{fig:network}
\end{figure}

An CSAM is a binary valued neural network in the sense that the weights on connections between a pair of neurons can either exist ($1$) or not ($0$).
Initially, no edge exists.
When storing a message, the corresponding active neurons and the connections in between are added into the network, forming a \emph{clique} (complete sub-graph). 
\fig{fig:network} depicts a toy CSAM with $C=5$ clusters and $L=4$ neurons each.
We label clusters and neurons clockwise.
Two messages are stored in this particular network instance, and the black clique expresses the message $(x_3, x_4, x_1, x_1, x_2)$.

Since a stored message corresponds to a clique in the network, it is equivalent to retrieve a message or its corresponding clique.
More precisely, when a probe is given (either a partial message $(?, x_4, x_1, ?, ?)$, or a corrupt one $(x_4, x_4, x_1, x_2, x_2)$), the network dynamic aims at converging to a clique containing similar neurons.
Previous studies~\cite{gripon2011sparse,gripon2012nearly,yao2013gpugbnn,yao2013bogus} focus on the erasure (partial message) scenario.
However, corrupt probes are more often in real world applications.
Think of digit images in USPS dataset, we are more likely to work with noisy images (black pixels becomes white or vice versa) than partial images.
Even if the erasure scenario is considered, it is often tricky to recognize where the missing pixels are in advance.
Unfortunately, this is a prerequisite for the existing retrieving algorithms \sos{}~\cite{gripon2011sparse} and \som{}~\cite{gripon2012nearly} to initialize correctly.
Therefore, both need to be revised to extend their use.

\subsection{Retrieving Algorithms}
Active neurons are considered as energy sources emitting signals along the edges.
Two basic iterative retrieving algorithms exist for CSAMs.
The set of active neurons at the end of the iterative retrieval process corresponds to the network response to the probe.
\subsubsection{\sos{}}
The default rule \sos{}~\cite{gripon2011simple,gripon2011sparse} for CSAMs is also applied in the model of Moopenn~\etal~\cite{moopenn1987electronic}.
Initially, neurons corresponding to the remaining probes are active, transmitting signals, whereas all neurons in the missing clusters deactivate.
After each iteration, neurons receive variable numbers of signals.
Only the neurons with the most signals in each cluster remain active in the next iteration.

Let $w_{(c,l)(c',l')}$ denote the indicator function of whether $\neuron(c,l)$ connects to $\neuron(c',l')$, i.e.,
\begin{equation}
    \label{eq:oldindex:w}
    w_{(c,l)(c',l')} =
    \begin{cases}
        1 & \mbox{$\neuron(c,l)$ connects to $\neuron(c',l')$}\\
        0 & \mbox{otherwise}.
    \end{cases}
\end{equation}
Let $v_{c,l}^t$ denote the indicator function of the potential for $\neuron(c,l)$ in iteration $t$, i.e.,
\begin{equation}
    \label{eq:oldindex:v}
    v_{c,l}^t =
    \begin{cases}
        1 & \quad\mbox{$\neuron(c,l)$ is active in iteration $t$}\\
        0 & \quad\mbox{otherwise}.
    \end{cases}
\end{equation}
We denote by $s_{c,l}^t$ the score, i.e., the count of the number of signals $\neuron(c,l)$ receives at iteration $t$.
The \sos{} retrieving algorithm are given by
\begin{align}
    s_{c,l}^t & = v_{c,l}^t + \sum_{c'=1}^{C}{\sum_{l'=1}^{L}{(v_{c',l'}^t w_{(c',l')(c,l)})}}\label{eq:oldindex:score}\\
    s_{c,\max}^t & = \max_{1\leq l\leq L}{s_{c,l}^t}\label{eq:findmax}\\
    v_{c,l}^{t+1} & =
    \begin{cases}
        1 & \quad\mbox{if $s_{c,l}^t = s_{c,\max}^t$}\label{eq:chooseMax}\\
        0 & \quad\mbox{otherwise}.
    \end{cases}
\end{align}
Essentially, \eq{eq:oldindex:score} counts the score for each neuron.
It involves summing over all clusters and all neurons within each cluster, hence the name \sos{}.
\eq{eq:findmax} finds the neurons with the most signal in each cluster, and \eq{eq:chooseMax} keeps them active.

\subsubsection{\som{}}
In contrast, \som{}~\cite{gripon2012nearly} activates a neuron if and only if it receives signals from every other clusters plus the self excitation.
Multiple signal contributions from the same cluster do not sum up.
However, in order for \som{} to succeed, it is prerequisite to activate all the neurons in the missing clusters initially, exactly opposite to \sos{}.

\som{} modifies the procedure to be
\begin{align}
    s_{c,l}^t &= v_{c,l}^t + \sum_{c'=1}^{C}{\max_{1\leq l'\leq L}{\left( v_{c',l'}^tw_{(c',l')(c,l)}\right)}}\label{eq:newrule:score}\\
    v_{c,l}^{t+1} &=
    \begin{cases}
        1 & \quad\mbox{if} \quad s_{c,l}^t =C\\
        0 & \quad\mbox{otherwise}.
    \end{cases}\label{eq:newrule:select}
\end{align}
The formulation above is essentially the same as in~\cite{gripon2011sparse,gripon2012nearly}, with the reinforcement factor $\gamma=1$ as suggested in~\cite{yao2013gpugbnn}.

\subsubsection{Advantages of \som{}}
Following the terminologies in~\cite{aboudib2013study}, \eq{eq:oldindex:score} and \eq{eq:newrule:score} are dynamic rules computing neuron potential scores, whereas \eq{eq:findmax}, \eq{eq:chooseMax} and \eq{eq:newrule:select} are activation rules determining neuron activities.
In other words, \sos{} counts individual signals as the dynamic rule and selects the local winners with the most signals as the activation rule; \som{} counts clusterwise signals as the dynamic rule and eliminates losers short of signals as the activation rule.

This iterative process continues until the network converges if it ever does.
As a matter of fact, we construct a counter example in~\cite{yao2013gpugbnn} showing that \sos{} might oscillate even in very basic cases.
In addition, we manage to prove the convergence of \som{}.
This is one of the reasons we prefer \som{} over \sos{}.
The second reason is that \sos{} counts individual signals possibly propagating decoding errors from iteration to iteration, whereas \som{} counts clusterwise contributions which stops signals from the same cluster summing up.
This explains why \som{} outperforms \sos{} in terms of retrieval rate by a large margin, especially in challenging scenarios, e.g., either the number of stored messages or the number of erased symbols increases.
For detailed performance comparisons and different initialization schemes, see~\cite{gripon2012nearly,yao2013gpugbnn}.
The last but not least reason is that \som{} sometimes leads to simpler hardware.
For instance, the authors in~\cite{jarollahi2013reduced} implement \som{} using AND and OR gates only, eliminating the requirement for inefficient comparison and addition units as in \sos{}.
%Here we illustrate another possible advantage to favor \som{}.
%Preferring clusterwise over individual signal counting not only brings us better retrieval rate~\cite{gripon2012nearly,yao2013gpugbnn}, but sometimes leads to simpler implementation as well.
%Although in~\cite{yao2013gpugbnn}, the nonlinear operations of \som{} runs much slower than the matrix-vector product of \sos{}, which is a highly optimized operation on GPUs, clusterwise signals turn out to be very efficient when implementing in logic circuits.
%Think of the network as a graph so that we have aggregated representations, i.e., the adjacency matrix $W$ of size $n\times n$ and the neuron vector $v$ of length $n$, where local index $(c,l)$ is replaced by the global index $k=(c-1)L+l$.
%To check if $\neuron(k)$ stays active or not in the next iteration, the $k$\textsuperscript{th} row (or column due to symmetry) of $W$ needs to be investigated. 
%\eq{eq:newrule:score} and \eq{eq:newrule:select} can be transformed to
%\begin{equation}
%	v_k^{t+1} = (W_{k,1}\wedge v_1^t\vee\cdots\vee W_{k,L}\wedge v_L^t)\wedge\cdots\wedge (W_{k,n-L+1}\wedge v_{n-L+1}^t\vee\cdots\vee W_{k,n}\wedge v_n^t),
%\end{equation}
%where each parenthesis represents a cluster.
%Unlike individual signals requiring compare and addition units, a neuron value in \som{} can be determined instantly using only $n$ AND gate with $2$ inputs, $1$ AND gate with $C$ inputs and $C$ OR gate with $L$ inputs.

\section{Source of Errors\label{sec:error}}
Let us analyze the source of errors which might influence the network.
We assume no error in the storing phase.
Associative memories can be viewed as a decoder in a communication system.
Due to the imperfect nature of communication channels, the receiver tries to recover a noisy signal from a predefined codebook (stored messages).
Therefore, a straightforward type of errors is the channel error, which in our context is a corrupt symbol in the probe.

Interestingly, another less obvious type of errors also exists, as we look closely into the way how CSAMs work.
Keep in mind that CSAMs do not manipulate symbols directly.
A message has to be first transformed to a sparse binary representation, upon which the network operates.
This encoding step itself and the following retrieving iterations might introduce errors as well.
We call this type the retrieving error.

Retrieving error may occur in three basic ways and the combination of them:
\begin{itemize}
\item Insertion. Multiple neurons activate in a cluster.
\item Omission. The desired neuron deactivates.
\item Shift. A spurious neuron activates in place of the expected one. We can also think of a shift being an omission followed by an insertion at a wrong position.
\end{itemize}
Although similar, an omission error is a more general but challenging situation than a partial probe in the erasure scenario.
The latter assumes both the number and the positions of the missing symbols are known in advance to the retrieving procedure.
This is not the case for omission errors.
It is the responsibility of the retrieving algorithm to make reasonable decisions on the fly.
Note that a shift error is indistinguishable from a channel error assuming the binary representations are flawless.
Therefore, if a retrieving algorithm covers all three aspects of retrieving errors, it should perform well against channel errors as well, and in both erasure and corruption scenarios.

\sos{} and \som{} behave differently towards the three aspects.
The differences root in the dynamic rules, i.e., counting the individual or clusterwise signals.
\sos{} erroneously accumulates neuron scores by the insertion errors, to which \som{} is innately immune.
This is exactly the reason that \som{} outperforms \sos{} by a large margin for the erasure scenario, as verified by the experiments in~\cite{yao2013gpugbnn}.
To confirm this claim, think of a loaded network after the storing phase.
Given a partial probe, during the initialization phase, the neurons in the missing clusters unconditionally deactivate, whereas the neurons corresponding to the remaining symbols activate.
After one iteration, all the cliques containing the remaining neurons are active, among which the desired message is included.
The missing clusters possibly contain multiple active neurons in this case, which is as if insertion errors happen.

\som{} is a monotonically decreasing procedure, in the sense that an inactive neuron can never revive back into play.
It performs poorly against omission and shift errors, because as soon as the remaining active neurons fail to form a clique during the retrieving process, all neurons throughout the network will deactivate immediately due to \eq{eq:newrule:select}.
This phenomenon is proved in~\cite{yao2013gpugbnn} and witnessed in~\cite{yao2013bogus} when applying heuristics to bypass the bogus fixed point.
To visualize such a scenario, let us revisit the black clique $(x_3, x_4, x_1, x_1, x_2)$ in~\fig{fig:network}, but with a corrupt probe $(x_2, x_4, x_1, x_1, x_2)$ presented.
It is easy to check that $\neuron(1,2)$ does not have any edge connecting to other neurons, thus cannot transmit signals.
Consequently, $\neuron(2,4)$, $\neuron(3,1)$, $\neuron(4,1)$ and $\neuron(5,2)$ do not receive any signal from cluster 1, and deactivate altogether.
This is very disappointing because a small perturbation can easily collapse the whole retrieval.

\section{Amendment to Structure\label{sec:amendment}}
Before making adaptations to the existing algorithms to cope with corrupt probes right away, we detour to make an amendment to the network structure.
Note that a probe might associate with multiple messages (cliques), then a natural question is which clique should be favored above the others during the retrieving process.
Previous work regards the number of a neuron's incident edges as a popularity measure of its counterpart symbol.
Throughout the stored messages, if $c$\textsuperscript{th} symbol often takes the $l$\textsuperscript{th} value, i.e., $m_c=x_l$, then $\neuron(c,l)$ tends to have more edges connected to it.
In~\cite{yao2013bogus}, the heuristics proposed to bypass the bogus fixed point depend implicitly on this measure.
When developing the clique finding algorithm later in the same work, we explicitly exploit this concept to determine the expanding order in the branch-and-bound recursive search, which brings us not only acceleration but also better retrieval rate.
However, this measure is biased in itself.
Since the binary weights on the edges $w_{(c,l)(c',l')}$ indicate either existence ($1$) or not ($0$), they leave us no clue whether an edge is shared by multiple messages.
This is a common situation, especially in a network with heavy loads.
The authors in~\cite{aboudib2013study} illustrate an example showing decoding failures in the case of overlapping edges.

In this work, we propose to fix this bias.
In addition to be aware of its incident edges, $\neuron(c,l)$ also keeps track of a frequency scalar $f_{c,l}$ during the storing phase, counting the number of times when $m_c=x_l$, i.e.,
\begin{equation}
f_{c,l} = \sum_{\text{all stored messages}}{\mathbbm{1}_{[m_c=x_l]}},
\end{equation}
where $\mathbbm{1}$ is the indicator function.

The simple construct introduces little overhead, because the number of additional variables grows linearly with the number of neurons in the network, not quadratically like the number of potential edges does.
To be precise, in a network of $C$ clusters with $L$ neurons, the $CL\times CL$ symmetric adjacency matrix is sufficient to describe the whole associate memory.
Therefore, $\binom{CL}{2}$ bits are consumed by the network itself.
The overhead introduced by the frequency is at most $CL\log_2{m}$ where $m$ is the number of stored messages.
It has been proved in~\cite{gripon2011sparse} that $m$ can not be proportional to $L^2$, so the overhead is neglectable.
Nevertheless, the retrieving process can benefit from this construct considerably.
One plausible way to use the frequency is to build preference or confidence if a probe retrieves multiple messages.
We can also construct normalized measures, e.g., average frequency per neuron in a retrieved clique, to extend CSAMs to store messages of variable lengths~\cite{aboudib2013study}.

\section{Retrieving Corrupt Messages\label{sec:proposal}} 
For description convenience, we should be acquainted with the aggregated representation $s$, $v$ and $W$.
Here $s$ and $v$ are vectors of length $n=CL$, respectively the stacked vectors of $s_{c,l}$ and $v_{c,l}$ by the cluster index, and $W$ is the $n\times n$ adjacency matrix describing all $w_{(c,l)(c',l')}$ with the diagonal elements being $1$. %, due to the reinforcement factor $\gamma=1$.
Colons (:) are seen occasionally in subscripts, indicating a particular index looping over all possible values.
For instance, $s_{c,:}$ stands for the scores of all neurons in cluster $c$, with the neuron index ranging from $1$ to $L$.
We will use all the notations whichever appears to be the most convenient to us.

For the completeness of our discussion, we also illustrate two building modules: \textbf{Joint}~\cite{yao2013gpugbnn} and \textbf{FindClique}~\cite{yao2013bogus} with small modifications in Appendix.
In brief, \textbf{Joint} is to apply one iteration of \sos{} followed by \som{}, an optimized way to retrieve the fixed point state after the network converges; \textbf{FindClique} is to find cliques in the fixed point state corresponding to stored messages.
As mentioned in Section~\ref{sec:amendment}, \textbf{FindClique} in the original work~\cite{yao2013bogus} determines the expanding order according to the number of a neuron's incident edges.
Since Section~\ref{sec:amendment} fixes this bias, we change the measure to the frequency vector $f$, the stacked version of all $f_{c,l}$.

\subsection{Direct Approach}
Applying standard \sos{} to the corruption scenario is able to produce meaningful retrievals only when the network is lightly loaded.
In this direction, \cite{aboudib2013study} discusses several more complicated variants, but they usually require user defined parameters which are not so obvious to determine\footnotemark.
\footnotetext{The network structure considered in~\cite{aboudib2013study} is not exactly the same. They consider messages of variable lengths, whereas we make full use of all the clusters.}

At this point, we argue that the performance of the direct approach degenerates drastically when the network becomes saturated.
Another huge problem with this approach is that it is not guaranteed to converge after a finite number of iterations thus some iteration cap is required to terminate the algorithm.

\subsection{Direct Plus Approach}
\begin{singlespace}
\begin{algorithm}
\caption{The direct plus approach of retrieving corrupt messages.}
\label{alg:directplus}
\DontPrintSemicolon
\SetDataSty{textit}
\SetKwFunction{clique}{FindClique}
\SetFuncSty{textbf}
\KwIn{The adjacency matrix $W$, the input probe $v$}
\KwOut{The retrieved clique $Q$}
\BlankLine\BlankLine\BlankLine
\Repeat{$v'=v$}{
	\tcp{Compute clusterwise signals.}
	\For{$k\gets 1\;\KwTo\;n$}{
		$s_k\gets 0$\;
		\ForEach{c}{
			\ForEach{l}{
				\If{$W_{k,(c-1)L+l}=1\;{\bf and}\;v_{(c-1)L+l}=1$}{
					$s_k\gets s_k+1$\;
					\textbf{break}\;
				}
			}
		}
	}
	\tcp{Preserve local winners into the next iteration.}
	\ForEach{$c$}{
		$s_{c,\max}\gets \max_{1\leq l\leq L}{s_{c,l}}$\;
		\ForEach{$l$}{
			$v'_{c,l}\gets 1\enskip\textbf{if}\enskip s_{c,l}=s_{c,\max}\enskip\textbf{else}\enskip 0$\;				
		}	
	}
}
\BlankLine\BlankLine\BlankLine
\tcp{A variant of the clique finding procedure in~\cite{yao2013bogus}, see Appendix.}
\tcp{The input graph $G(V,E)$ is determined by $W$ and $v$, possibly a bogus fixed point.}
split $V$ into $C$ sets $R=\lbrace R_0,R_1,\cdots,R_{C-1}\rbrace$ with each accommodating the neurons in a different cluster\;
$Q\gets\clique{R}$\;
\end{algorithm}
\end{singlespace}

The direct approach of applying \sos{} is not a satisfactory solution to the corruption scenario.
We see in Section~\ref{sec:error} that counting individual signals makes \sos{} sensitive to insertions.
Although \som{} is not affected by insertion errors because its dynamic rule counts clusterwise signals, naively applying it does not work well against omission and shift errors due to its stringent activation rule, also mentioned in Section~\ref{sec:error}.
Therefore, a possible scheme is to hybrid clusterwise signals from \som{} with the activation rule of \sos{} (winner-takes-all) in Algorithm~\ref{alg:directplus}.
We call it the direct plus approach.
Two features are different from the direct approach:
\begin{itemize}
	\item Clusterwise signals replace individual signals.
	\item A clique finding procedure is applied at the end of the algorithm.
\end{itemize}
Although not completely, these two modifications do mitigate the convergence problem and change the characteristics of the direct approach, making the direct plus approach a strong competitor in the comparison; see Section~\ref{sec:experiment}.
%Mixing of \sos{} and \som{} can reduce the running time significantly~\cite{yao2013gpugbnn}, can help to escape from bogus fixed points~\cite{yao2013bogus}, improving the retrieval rate tremendously.

\subsection{Construct Approach}
\begin{singlespace}
\begin{algorithm}
\caption{The construct approach of retrieving corrupt messages.}
\label{alg:construct}
\DontPrintSemicolon
\SetKwFunction{clique}{FindClique}
\SetFuncSty{textbf}
\SetDataSty{textit}
\SetKwData{signal}{signal}

\KwIn{The adjacency matrix $W$, the input probe $v$}
\KwOut{The retrieved clique $Q$}
\BlankLine\BlankLine\BlankLine
\Repeat{$Q\neq\emptyset$}{
	\tcp{Compute clusterwise signals.}
	\For{$k\gets 1\;\KwTo\;n$}{
		$s_k\gets 0$\;
		\ForEach{c}{
			\ForEach{l}{
				\If{$W_{k,(c-1)L+l}=1\;{\bf and}\;v_{(c-1)L+l}=1$}{
					$s_k\gets s_k+1$\;
					\textbf{break}\;
				}
			}
		}
	}
	\BlankLine\BlankLine\BlankLine
	\tcp{Update the effective adjacency matrix $W'$.}
	\For{$k\gets 1\;\KwTo\;n$}{
		\For{$k'\gets 1\;\KwTo\;n$}{
			\If{$W_{k,k'}=1\;{\bf and}\;v_k=1\;{\bf and}\;v_{k'}=1$}{
				$W'_{k,k'}\gets 1$
			}
		}
	}
	\BlankLine\BlankLine\BlankLine
	\tcp{A variant of the clique finding procedure in~\cite{yao2013bogus}, see Appendix.}
	\tcp{The input graph $G(V,E)$ is determined by $W'$ and $V$.}
	obtain a smaller graph $G'(V',E')$ that only contains neurons with their clusterwise signals being exactly $C$\;
	split $V'$ into $C$ sets $R=\lbrace R_0,R_1,\cdots,R_{C-1}\rbrace$ with each accommodating the neurons in a different cluster\;
	$Q\gets\clique{R}$\;
	\BlankLine\BlankLine\BlankLine
	\tcp{Set an inactive neuron with the most clusterwise signals active.}
	$a\gets\argmax_{k\in\lbrace v_k=0\rbrace}{\lbrace s_k\rbrace}$\;
	$v_a=1$\;
}
\end{algorithm}
\end{singlespace}

As mentioned before, \som{} is a monotonically decreasing procedure.
It finds a superset of neurons that includes the desired message and eliminates improbable neurons gradually.
We propose to go along the opposite direction in Algorithm~\ref{alg:construct}, adding probable neurons into a subset of neurons gradually and constructing the cliques from scratch.
Instead of having the clique finding procedure at the end as a post-processing step in~\cite{yao2013bogus}, \textbf{FindClique} is now served as an integrated part of the algorithm.
The basic idea is to think of all active neurons comprising an active set.
Every iteration, we add some neurons into this active set, and try to find a clique within it.
The retrieving process terminates once a clique is encountered.
Note that \textbf{FindClique} is modified slightly from \cite{yao2013bogus} in that only the neurons with their clusterwise signals being exactly $C$ can go into the recursive procedure.
This condition is not required there in~\cite{yao2013bogus} as it is ensured by \som{} implicitly.

Line 2--12 compute the clusterwise signal for each neuron.
Line 13--19 update the effective adjacency matrix, since not all the edge weights in the original adjacency matrix $W$ are relevant to the active set thus to the clique finding procedure.
Line 23--24 look through all the inactive neurons and add one with the most signals into the active set.

Unfortunately, the plain construct approach in Algorithm~\ref{alg:construct} runs discouragingly slow due to nested levels of loops.
Three tricks can help to reduce the running time significantly.
The first trick lies when we compute the clusterwise signals at line 2--12.
Since the score of each neuron increases monotonically with the number of active neurons, we do not need to update the score function for every neuron.
Instead, we only update the neurons whose clusterwise signal is less than the number of clusters $C$.

Notice that updating the effective adjacency matrix at line 13--19 inside the big loop is a time consuming operation even in a network of a reasonable size, since the adjacency matrix is quadratic in the number of neurons.
In fact, it is not necessary to update the whole adjacency matrix.
The active set enlarges incrementally, which means all neurons in the current iteration still exist in the next.
Therefore, only the elements associated with the newly added neuron need to be updated, i.e., if $\neuron(c,l)$ is added, only the $(c-1)L+l$\textsuperscript{th} row (and column) of the adjacency matrix needs to be updated.
This modification brings the time complexity down from $O(n^2)$ to $O(n)$.

The third trick is in line 23--24, rather than adding one neuron with the most clusterwise signals into the active set, we subjoin multiple neurons at a time in the hope of finding a clique in earlier iterations and quitting the algorithm as soon as possible.
There are plenty of subjoining schemes that could be applied, which balance between retrieval rate and running time.
For instance, we could add all neurons with the most clusterwise signals into the active set at one shot, or we could look into different clusters and add the neurons with the most signals in each cluster.
The more neurons are added each time, the sooner the algorithm can find a clique, and the shorter the running time will be.
However, adding too many neurons potentially increase the number of cliques that can be found in the network; the probability of retrieving the correct one decreases.
Due to good empirical performance, we propose to check through all inactive neurons, subjoining it if its score is not lower than all the previous neurons.

\subsection{Delegate Approach}
\begin{singlespace}
\begin{algorithm}
\caption{The delegate approach of retrieving corrupt messages.}
\label{alg:delegate}
\DontPrintSemicolon
\SetDataSty{textit}
\SetKwFunction{clique}{FindClique}
\SetKwFunction{joint}{Joint}
\SetFuncSty{textbf}
\KwIn{The adjacency matrix $W$, the input probe $v$}
\KwOut{The retrieved clique $Q$}
\BlankLine\BlankLine\BlankLine
\tcp{Convert corruption scenario into erasure case.}
\tcp{Checking clusterwise signals is also feasible, resilient to insertions.}
$P\gets\emptyset$\tcp*[f]{The set of missing clusters.}\;
$s\gets W\cdot v$\tcp*[f]{Can be replaced by clusterwise signals.}\;
\ForEach{$c$}{
	$s_{c,\max}\gets \max_{1\leq l\leq L}{s_{c,l}}$\;
	\ForEach{$l$}{
		$v'_{c,l}\gets 1\enskip\textbf{if}\enskip s_{c,l}=s_{c,\max}\enskip\textbf{else}\enskip 0$\;
	}
	\If{$v'_{c,:}\neq v_{c,:}$}{
		$v_{c,:}\gets 0$\;
		$P\gets P\cup\{c\}$
	}
}
\BlankLine\BlankLine\BlankLine
\tcp{The joint scheme in~\cite{yao2013gpugbnn}, see Appendix.}
\tcp{At this point, we know the positions $P$ of the missing clusters.}
\joint{$W, v, P$}
\BlankLine\BlankLine\BlankLine
\tcp{A variant of the clique finding procedure in~\cite{yao2013bogus}, see Appendix.}
\tcp{The input graph $G(V,E)$ is determined by $W$ and $v$, possibly a bogus fixed point.}
obtain a smaller graph $G'(V',E')$ with $C'$ clusters by eliminating non-erased clusters and erased clusters but with only one active neuron\;
obtain an even smaller graph $\widetilde{G}(\widetilde{V},\widetilde{E})$ with $\widetilde{C}$ clusters by eliminating inactive neurons in the remaining clusters\;
split $\widetilde{V}$ into $\widetilde{C}$ sets $R=\lbrace R_0,R_1,\cdots,R_{\widetilde{C}-1}\rbrace$ with each accommodating the neurons in a different cluster\;
$Q\gets\clique{R}$\;
\end{algorithm}
\end{singlespace}

The partial probe problem in erasure scenarios has already been solved elegantly by the joint scheme~\cite{yao2013gpugbnn} followed by the clique finding procedure~\cite{yao2013bogus}.
Therefore, if somehow we can transform the corrupt probe problem into the partial probe problem, \textbf{Joint} and \textbf{FindClique} can be exploited untouched.
We propose such a delegate approach illustrated in Algorithm~\ref{alg:delegate}.

In Algorithm~\ref{alg:delegate}, line 2, 4, 6 are equivalent to \eq{eq:oldindex:score}, \eq{eq:findmax} and \eq{eq:chooseMax} respectively.
Therefore, Algorithm~\ref{alg:delegate} essentially runs one iteration of \sos{} at the very beginning.
Line 8--11 check if the result coincides with the probe.
If the result is identical with the probe for some symbols, the algorithm builds up confidence about the correctness of these symbols.
The different clusters are then treated as missing symbols in the erasure scenario, relying on \textbf{Joint} and \textbf{FindClique} to rescue. 
Note that checking individual signals at line 2 can also be replace by clusterwise signals instead, which is believed resilient to insertion errors.

\subsection{Cut-and-paste Approach}
\begin{singlespace}
\begin{algorithm*}
\caption{The cut-and-paste approach of retrieving corrupt messages.}
\label{alg:cutandpaste}
\DontPrintSemicolon
\SetDataSty{textit}
\SetKwData{connected}{connected}
\SetKwFunction{clique}{FindClique}
\SetKwFunction{joint}{Joint}
\SetFuncSty{textbf}
\KwIn{The adjacency matrix $W$, the input probe $v$}
\KwOut{The retrieved clique $Q$}
\BlankLine\BlankLine
\tcp{Search for all cliques of different sizes in the probe by incrementally enlarging the targeted cluster.}
$A_1\gets\emptyset$\tcp*[f]{A set containing the active neurons of the probe.}\;
$A\gets\emptyset$\tcp*[f]{A set of sets containing all cliques that can be found in the probe.}\;
\For{$k\gets 1\;\KwTo\;n$}{
	\If{$v_k=1$}{
		$A_1\gets A_1\cup\{k\}$\;
		$A\gets A\cup\lbrace{\{k\}}\rbrace$\;
	}
}
\ForEach{$a\in A$}{
	\ForEach{$a_1\in A_1$}{
		$\connected\gets\textbf{true}$\;
		\ForEach{$a'\in a$}{
			\If{$W_{a_1,a'}\neq 1$}{
				$\connected\gets\textbf{false}$\;
				\textbf{break}\;
			}	
		}
		\If{$\connected$}{
			$A\gets A\cup\{a\cup\{a_1\}\}$\;		
		}
	}
}
sort $A$ according to clique sizes and the sum of the neuron frequencies if two cliques are of the same size\;
\BlankLine\BlankLine
\Repeat{$\card{Q}\neq 0$}{
	$P\gets\{1,2,\cdots,C\}$\tcp*[f]{The positions of the missing clusters.}\;
	$a\gets A[1]$\;
	$A\gets A\setminus\{a\}$\;
	$v\gets 0$\;
	\ForEach{$a'\in a$}{
		$v_{a'}\gets 1$\;
		$P\gets P\setminus\{\floor{\frac{a'-1}{L}+1}\}$\;
	}
	$s\gets W\cdot v$\tcp*[f]{Can be replaced by clusterwise signals.}\;
	\ForEach{$p\in P$}{
		\ForEach{$l$}{
			$v_{(p-1)L+l}\gets 1\enskip\textbf{if}\enskip s_{(p-1)L+l}=C-\card{P}\enskip\textbf{else}\enskip 0$		
		}
	}
	\tcp{A variant of the clique finding procedure in~\cite{yao2013bogus}, see Appendix.}
	\tcp{The input graph $G(V,E)$ is determined by $W$ and $v$, possibly a bogus fixed point.}
	split $V$ into $C$ sets $R=\lbrace R_0,R_1,\cdots,R_{C-1}\rbrace$ with each accommodating the neurons in a different cluster\;
	$Q\gets\clique{R}$\;
}
\end{algorithm*}
\end{singlespace}

The delegate approach smartly converts a more difficult problem into a known one and solved by existing techniques.
However, it comes with its own problem.
After the first iteration, clusters different from the probe are assumed to be missing symbols.
This operation does not ensure the remaining ``confident'' symbols can actually form a clique.
It is possible that no clique is found at the end of the algorithm, thus the retrieved ``message'' might never be stored in the network.
 
The cut-and-paste approach described in Algorithm~\ref{alg:cutandpaste}, although resembles the delegate approach, tries to solve this particular flaw.
Initially, we \emph{cut} out all cliques (not necessarily of size $C$) that can be found in the probe, and sort them in descending order according to the clique size and node frequency if multiple cliques are of the same size (line 1--23).
Then we \emph{paste} them one at a time back into the network, assuming the rest clusters are missing, and resort to \textbf{Joint} and \textbf{FindClique} (line 24--41).
The purpose of the cut stage is to make sure the remaining confident symbols are always a partial stored message, thus fixing the problem of the delegate approach.

It is necessary to explicitly differentiate the cliques found in the cut stage and the cliques reported as retrieved messages.
The cliques found in the cut stage, different in size, populate a candidate set.
Each member (clique) is assumed to be the confident symbols in the erasure scenario.
Therefore, it is not difficult to understand the rationale that the clique size and node frequency should be used to sort the candidates.
A larger clique encourages the retrieved message to be similar with the probe, since more symbols are assumed to be confident.
A higher node frequency increases the probability of reporting the correct message if two cliques are of the same size.
Keep in mind that finding \emph{all} cliques in the cut stage is essential to the success of this approach, since a clique might contain erroneous neurons, enlarging the clique size solely by chance and might not correspond to any stored message.

To better understand the cut-and-paste approach, let us revisit again the toy network in \fig{fig:network}.
A corrupt probe is given as $(x_1, x_1, x_1, x_1, x_2)$.
We first find out all cliques residing in the probe.
In this example, five cliques exist $(x_1, x_1, ?, ?, ?)$, $(?, ?, x_1, x_1, ?)$, $(?, ?, x_1, ?, x_2)$, $(?, ?, ?, x_1, x_2)$ and $(?, ?, x_1, x_1, x_2)$.
We sort them in descending order according to clique size, so the largest clique $(?, ?, x_1, x_1, x_2)$ will be favored.
We then consider it as the partial probe in an erasure scenario, the original message $(x_3, x_4, x_1, x_1, x_2)$ can be successfully retrieved.
Hypothetically, if the largest clique does not retrieve a stored message, then one of the smaller cliques will be tested.
Among the four cliques of the same size, the frequency scalars will be checked to determine the priority.

One maybe hesitate to search for all cliques in a probe because it is NP-hard.
However, this is not completely a disaster for cut-and-paste.
Qualitatively speaking, because of the sparsity that each cluster ideally accommodate one active neuron, the running time grows mildly with the number of clusters $C$.
Moreover, since CSAMs are $C$-partite graphs, the maximum value of the clusterwise signals is an upper bound of all possible clique sizes.
In addition, the upper bound is tight if no insertion error occurs.

\section{Experiments\label{sec:experiment}}
All the simulations are performed on a 2.4GHz Intel Q6600 with 8GB memory.

\subsection{Simulated Data}
We compare five approaches proposed in Section~\ref{sec:error} under different types of errors separately to better understand the pros and cons of each of them.
We simulate on $C=8$ clusters, $L=128$ neurons each, with increasing number of stored messages.
2000 messages are used for testing. %, and we run experiments 10 times to average out the final results.
Note that the running time for \sos{} is topped by a cap of $10$ iterations.
Otherwise, it may run arbitrarily slow due to its convergence problem.
%To make a fair comparison, we also include the celebrated Willshaw network as a baseline test, since most of the associative memory systems are built upon it.

\subsubsection{Insertion Errors}
\begin{figure*}
\centering
\subfloat[Few Light Insertion Errors]{
\includegraphics[scale=0.45]{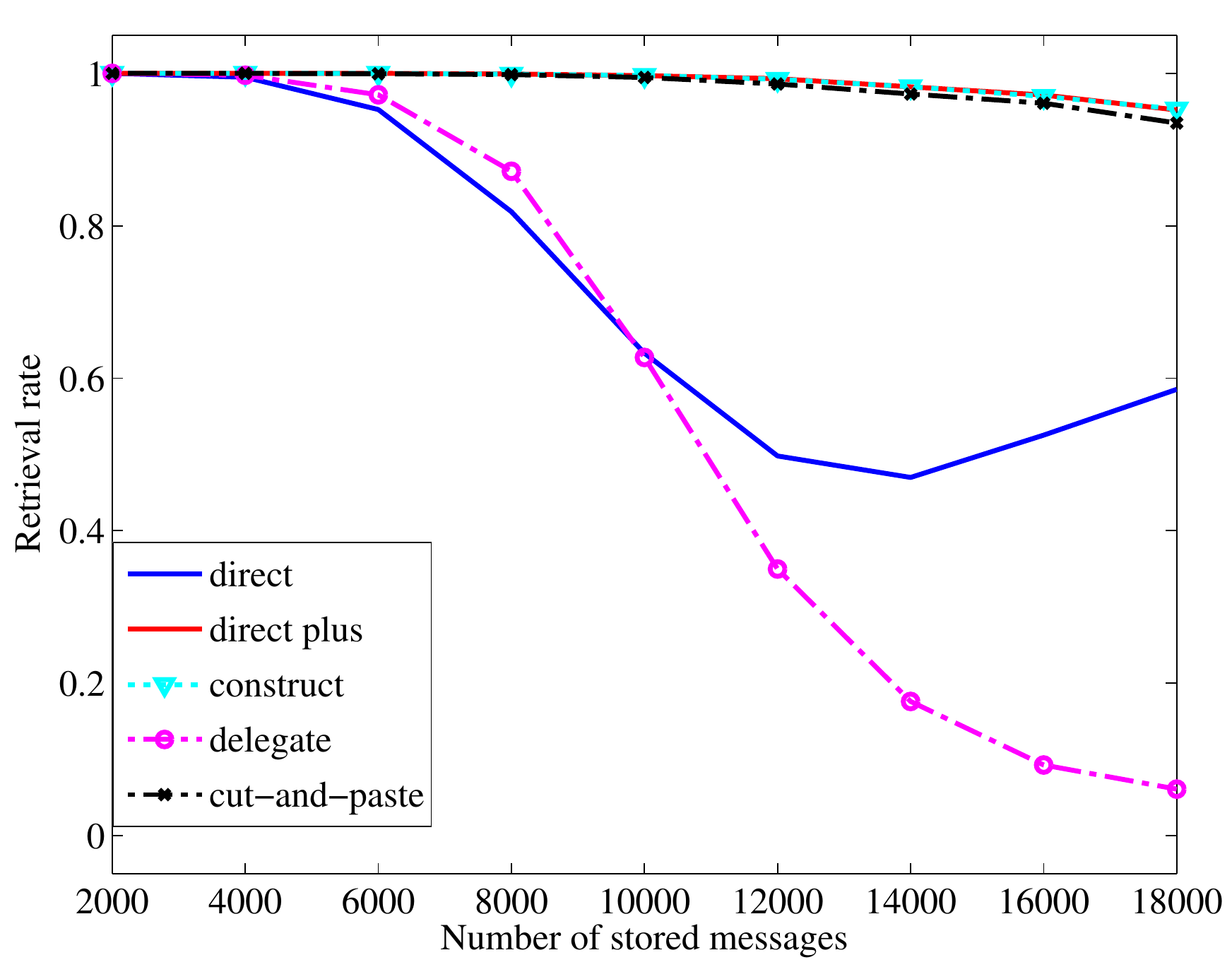}
\label{fig:fewlightrate}
}
\subfloat[Few Light Insertion Errors]{
\includegraphics[scale=0.45]{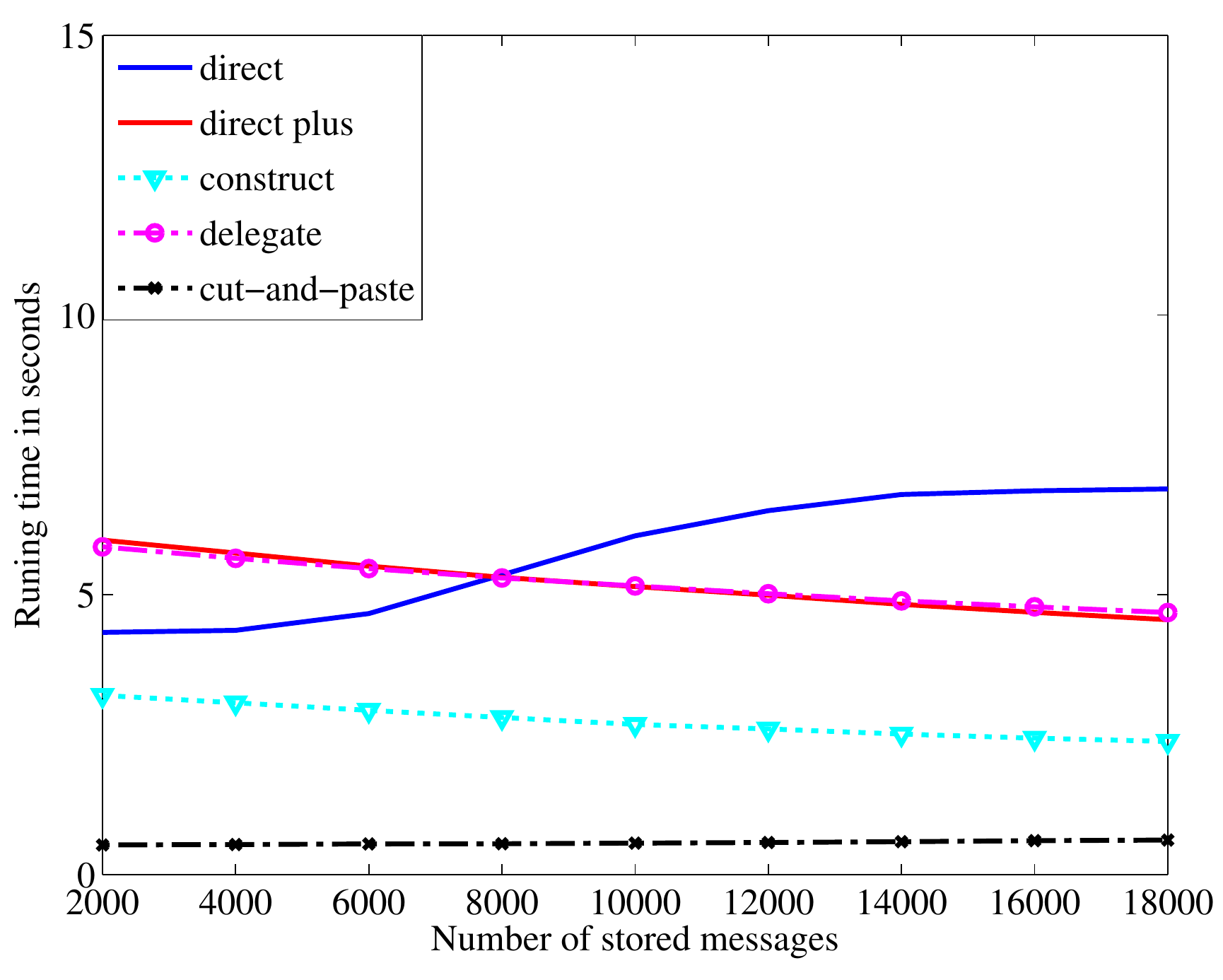}
\label{fig:fewlighttime}
}\\
\subfloat[Many Light Insertion Errors]{
\includegraphics[scale=0.45]{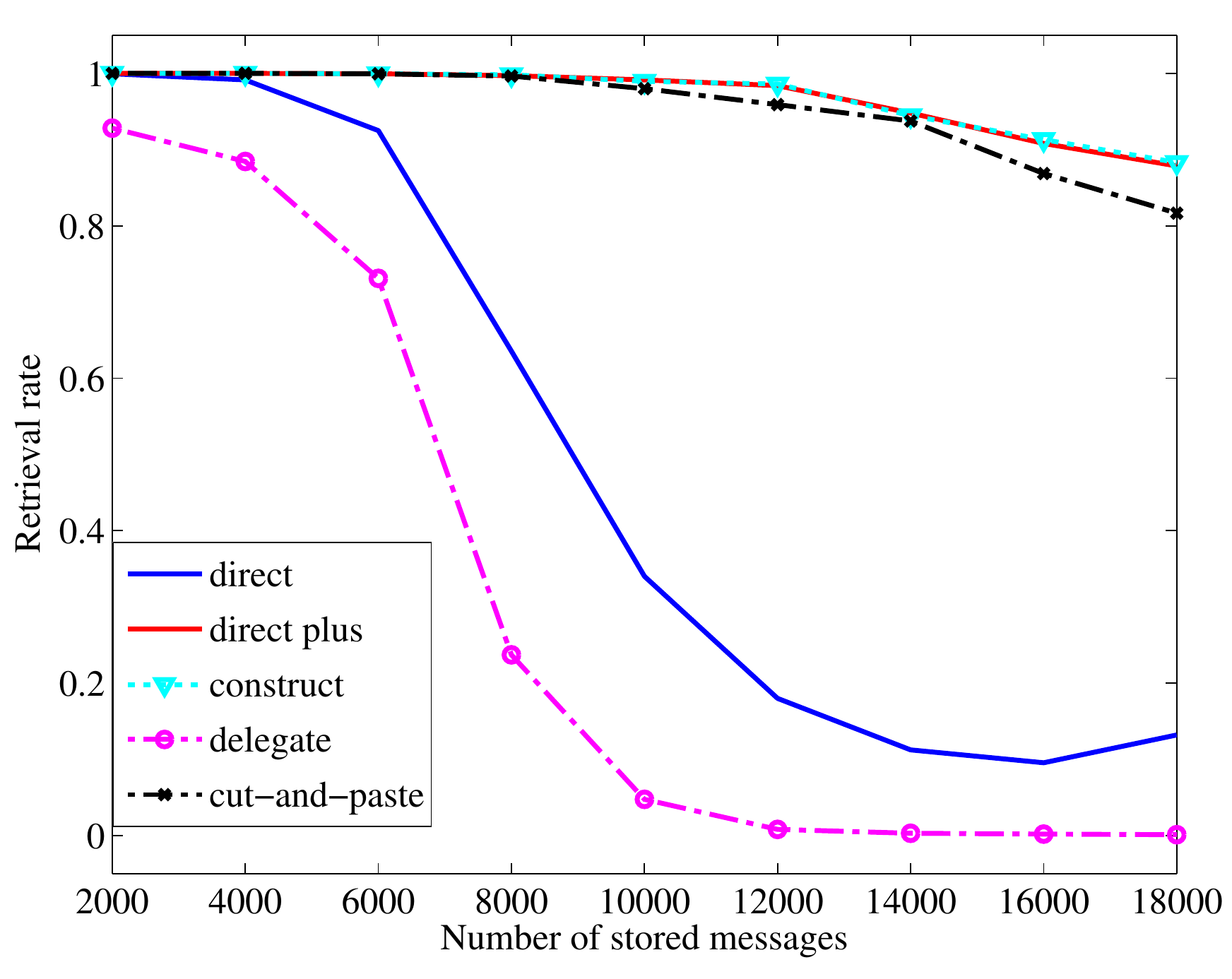}
\label{fig:manylightrate}
}
\subfloat[Many Light Insertion Errors]{
\includegraphics[scale=0.45]{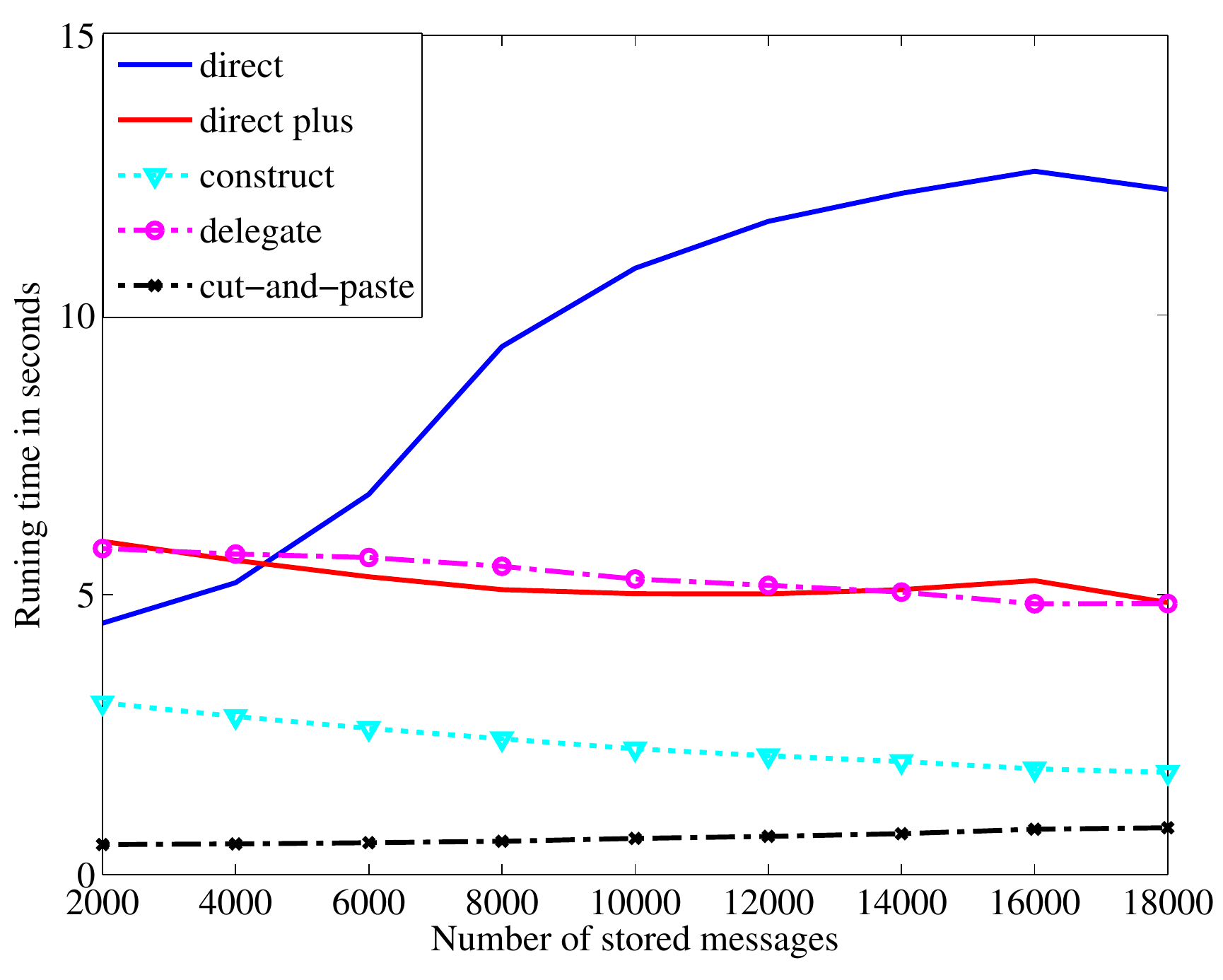}
\label{fig:manylighttime}
}\\
\subfloat[Few Heavy Insertion Errors]{
\includegraphics[scale=0.45]{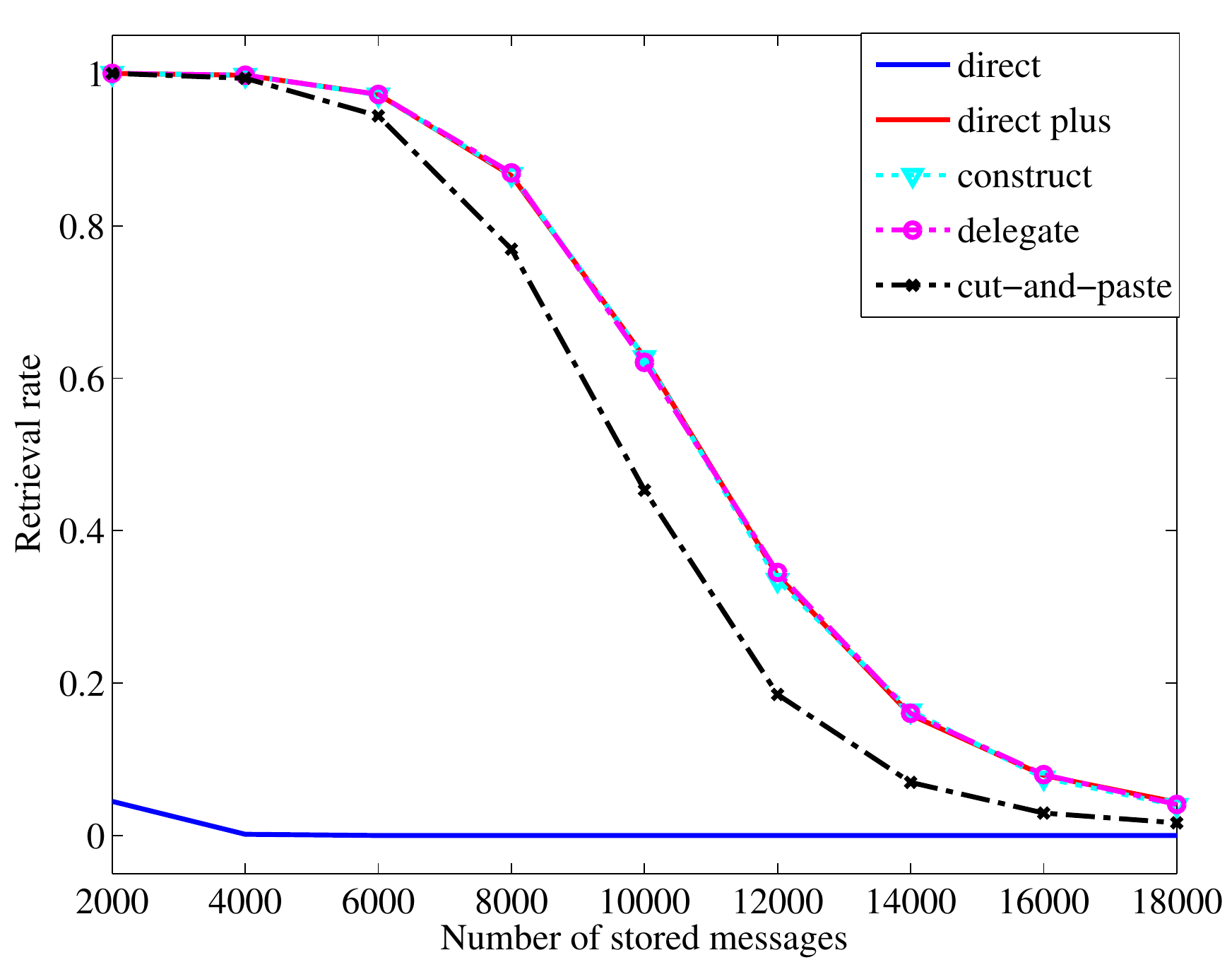}
\label{fig:fewheavyrate}
}
\subfloat[Few Heavy Insertion Errors]{
\includegraphics[scale=0.45]{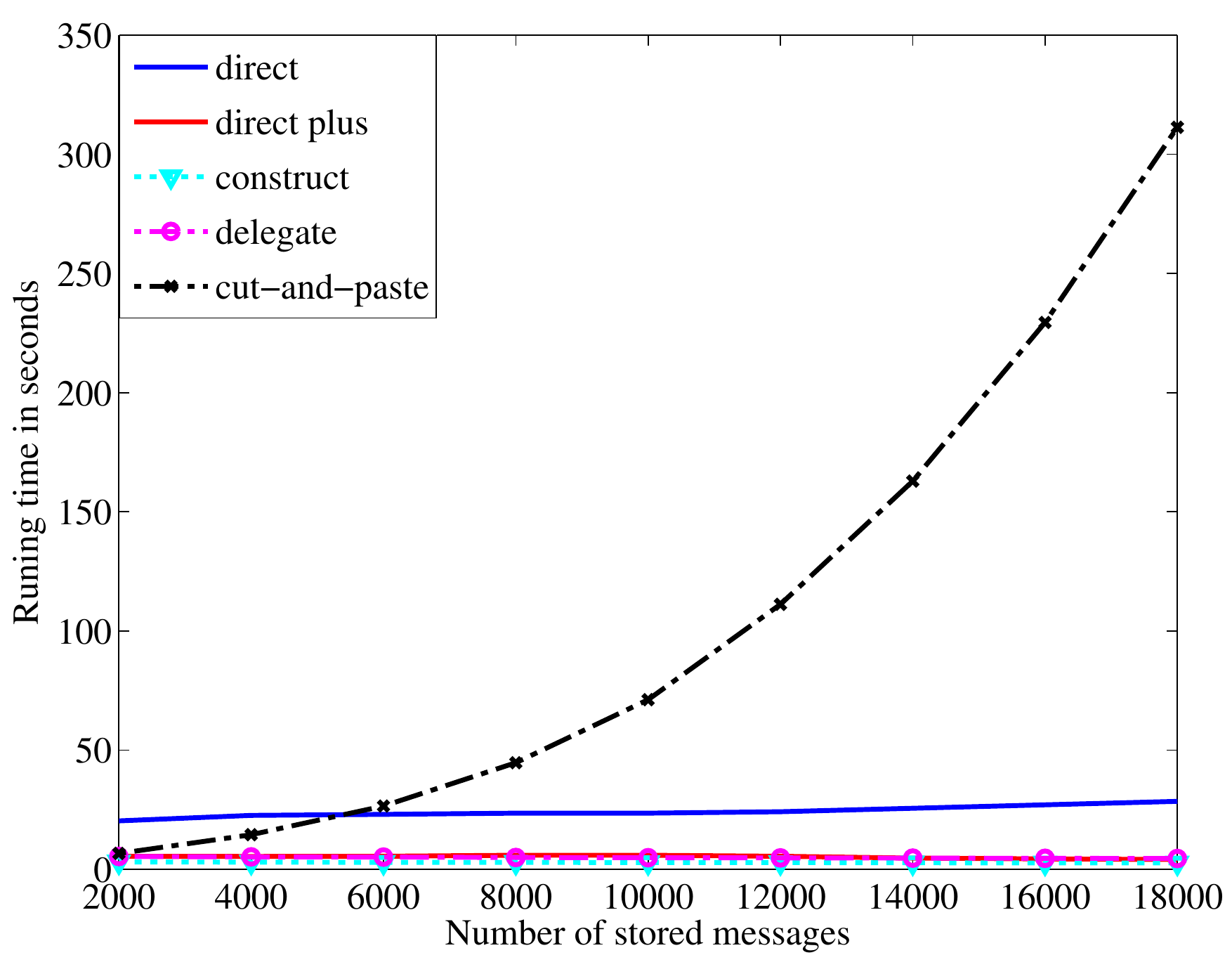}
\label{fig:fewheavytime}
}
\caption{Comparisons of different approaches proposed in Section~\ref{sec:error} under insertion errors. For \protect\subref{fig:fewlightrate} and \protect\subref{fig:fewlighttime}, the first neuron of two clusters are activated. For \protect\subref{fig:manylightrate} and \protect\subref{fig:manylighttime}, the first neuron of all clusters are activated. For \protect\subref{fig:fewheavyrate} and \protect\subref{fig:fewheavytime}, all neurons of two clusters are activated.}
\label{fig:insertion}
\end{figure*}

Insertion errors can happen in two distinct ways: (1) A large number of symbols suffer from light insertion errors (e.g., one insertion for each cluster). (2) One or two symbols suffer from heavy insertion errors (e.g., all neurons in a cluster are activated).
We show the comparison in \fig{fig:insertion}.
For \fig{fig:fewlightrate} and \fig{fig:fewlighttime}, the first neuron of two clusters are activated unconditionally. For \fig{fig:manylightrate} and \fig{fig:manylighttime}, the first neuron of all clusters are activated. For \fig{fig:fewheavyrate} and \fig{fig:fewheavytime}, all neurons of two clusters are activated.

We see from \fig{fig:fewlightrate} and \fig{fig:manylightrate} that the direct plus, construct and cut-and-paste approach perform extremely well, showing great advantages over the direct and delegate approach.
In terms of running time, the cut-and-paste is the biggest winner, which runs orders of magnitude faster.
The interesting part is in the case where few symbols are experiencing heavy load insertion errors as shown in \fig{fig:fewheavyrate} and \fig{fig:fewheavytime}.
This is the only case where cut-and-paste loses the competition.
The direct approach stops retrieving anything useful.
The cut-and-paste approach falls behind the other three alternatives and it loses its advantage in quick running time completely.

\subsubsection{Omission Errors}
\begin{figure*}
\centering
\subfloat[Few Omission Errors]{
\includegraphics[scale=0.45]{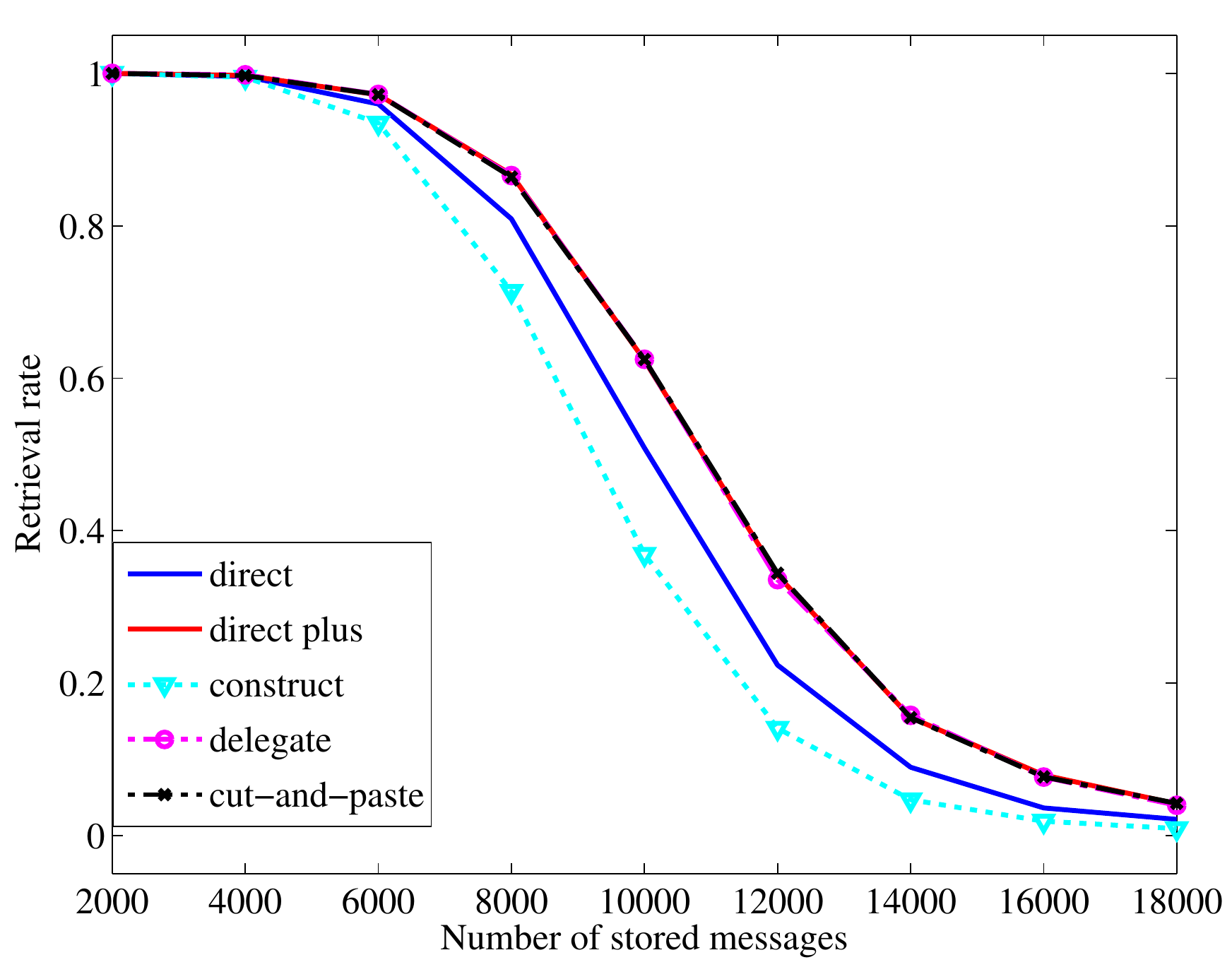}
\label{fig:fewomissionrate}
}
\subfloat[Few Omission Errors]{
\includegraphics[scale=0.45]{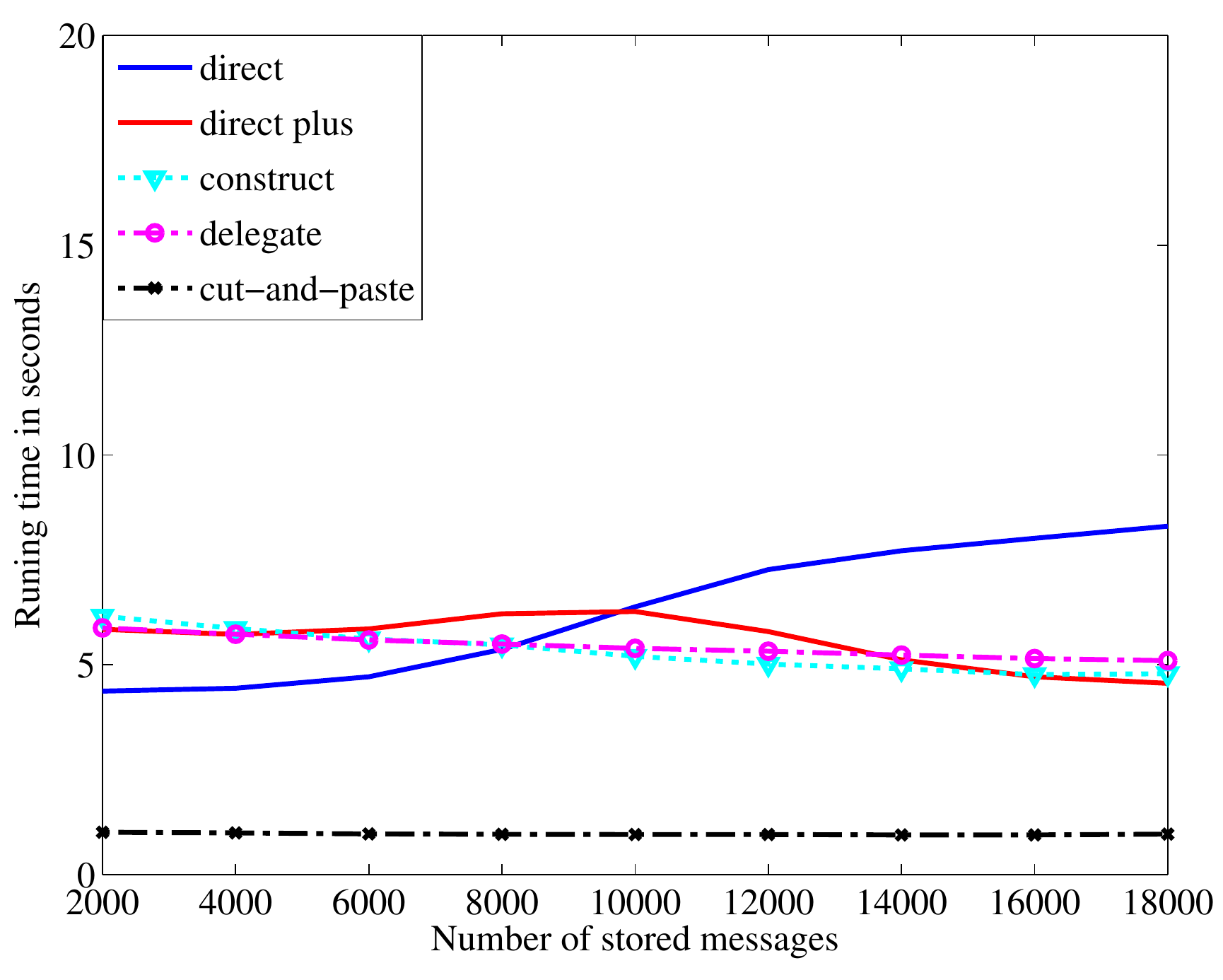}
\label{fig:fewomissiontime}
}
\\
\subfloat[Many Omission Errors]{
\includegraphics[scale=0.45]{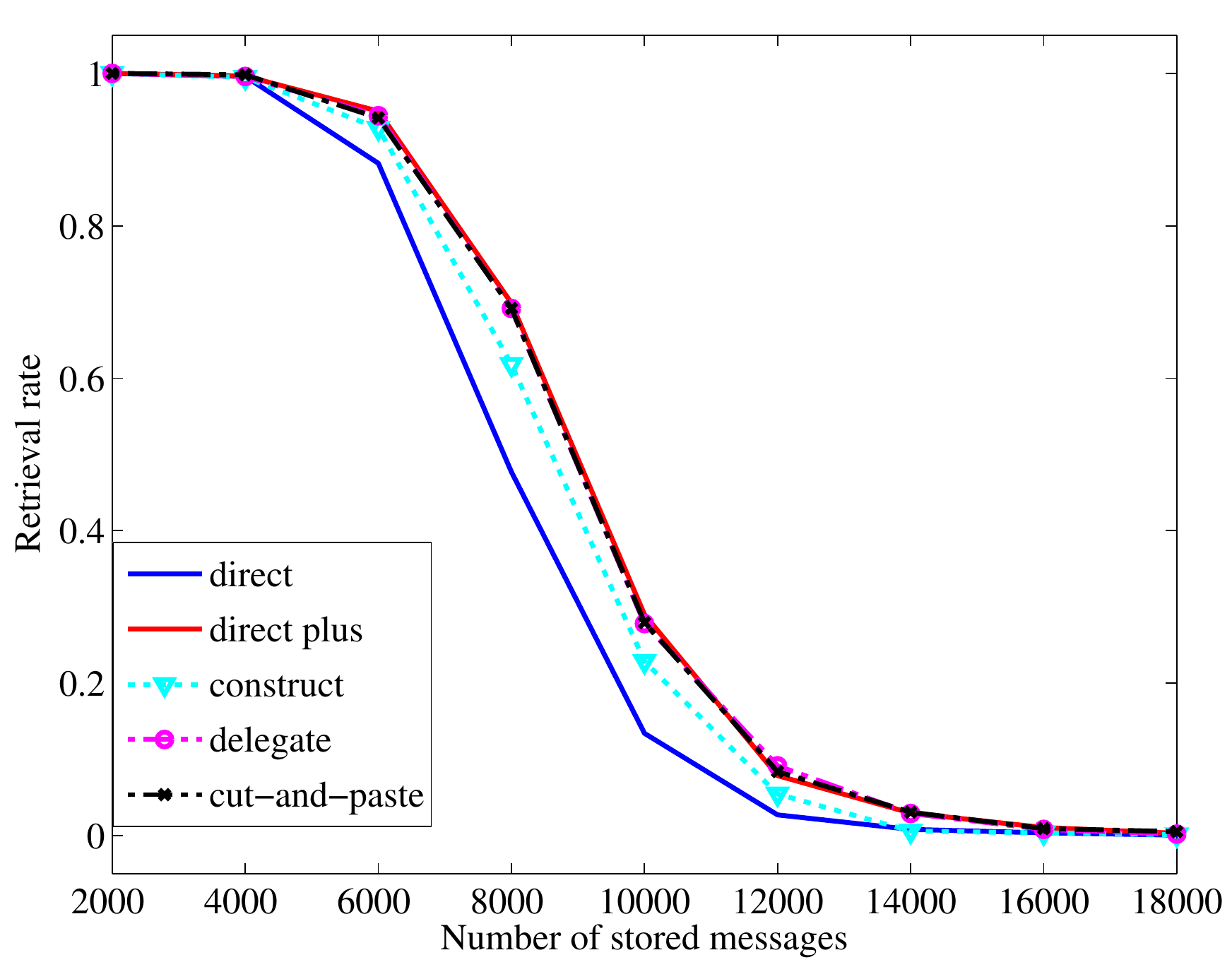}
\label{fig:manyomissionrate}
}
\subfloat[Many Omission Errors]{
\includegraphics[scale=0.45]{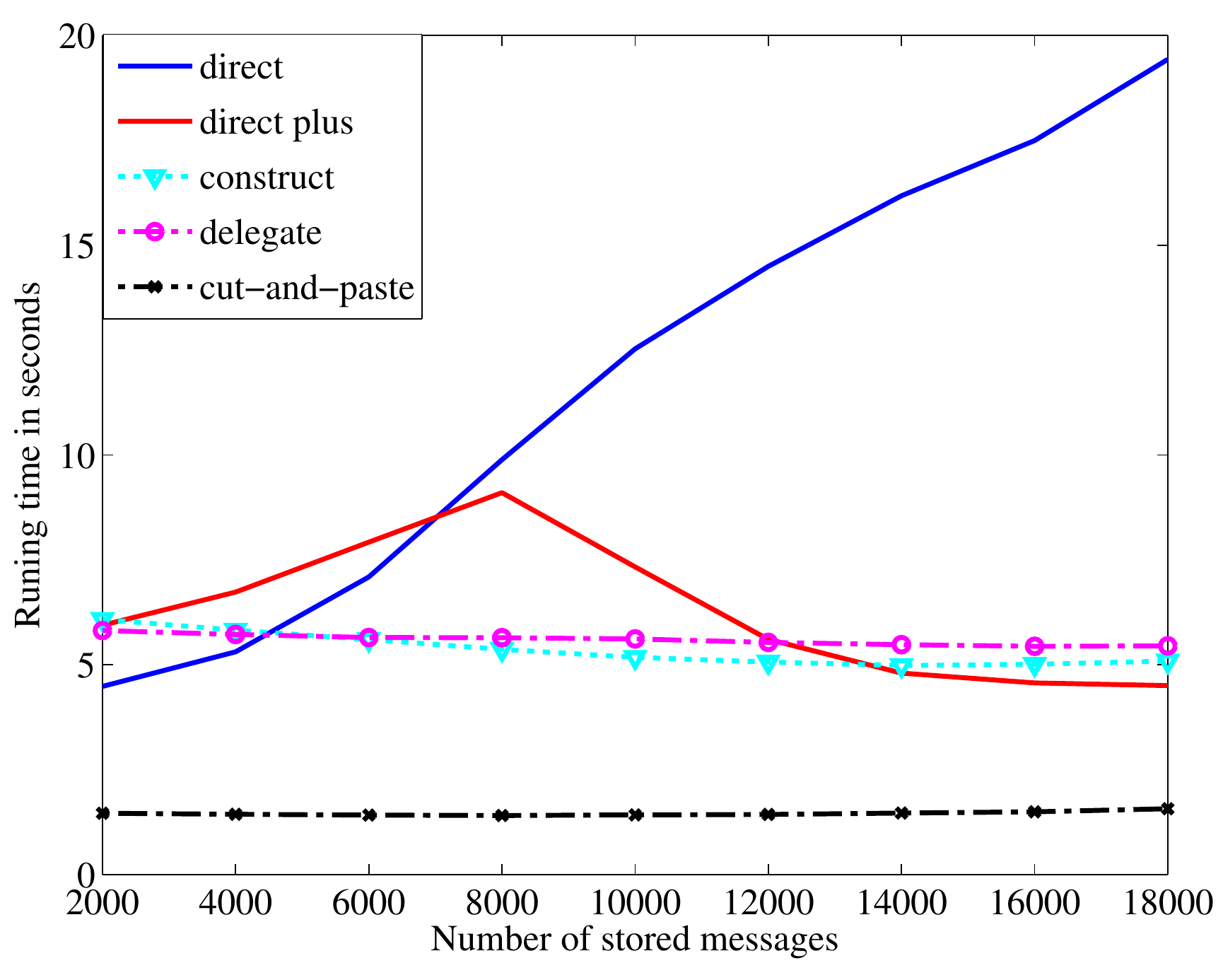}
\label{fig:manyomissiontime}
}
\caption{Comparisons of different approaches proposed in Section~\ref{sec:error} under omission errors. For \protect\subref{fig:fewomissionrate} and \protect\subref{fig:fewomissiontime}, two symbols suffer from omission errors. For \protect\subref{fig:manyomissionrate} and \protect\subref{fig:manyomissiontime}, four symbols are affected.}
\label{fig:omission}
\end{figure*}

Omission errors are equivalent to the erasure scenario without knowing the locations of the missing symbols in advance, which the algorithms are required to determine on the fly.
For \fig{fig:fewomissionrate} and \fig{fig:fewomissiontime}, two symbols suffer from omission errors.
For \fig{fig:manyomissionrate} and \fig{fig:manyomissiontime}, four symbols are affected.

We see from \fig{fig:omission} that the direct plus, delegate and cut-and-paste approach run almost identically, outperforming the direct and construct approach.
As omission errors increase, the direct approach is the most sensitive one so that it performs the worst in \fig{fig:manyomissionrate}.
In terms of running time, the cut-and-paste approach beats every other approaches by a large margin as always.
The running time of the direct approach seems linearly growing with the number of stored messages.

\subsubsection{Shift Errors}
\begin{figure*}
\centering
\subfloat[Few Low Probability Shift Errors]{
\includegraphics[scale=0.45]{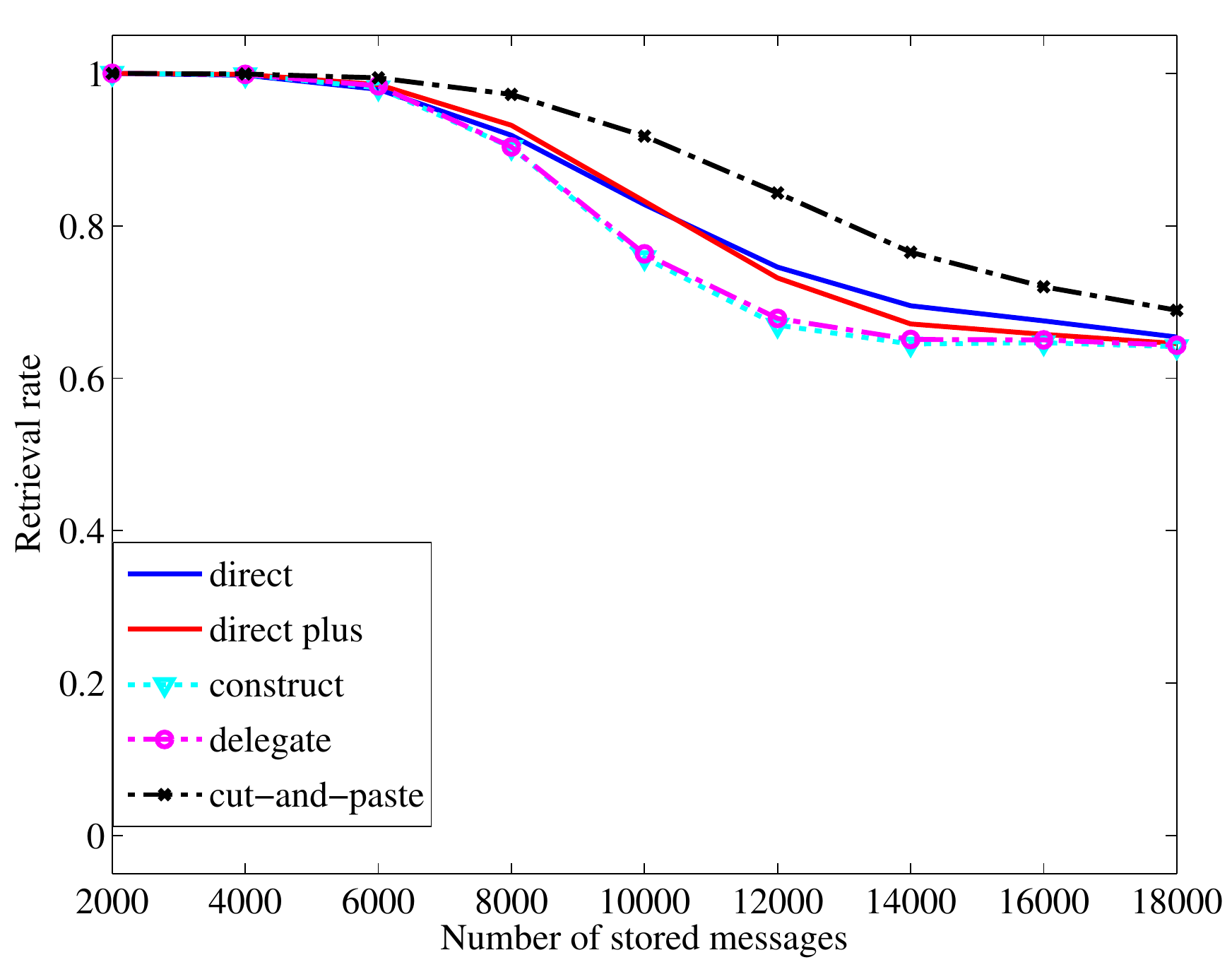}
\label{fig:fewlowrate}
}
\subfloat[Few Low Probability Shift Errors]{
\includegraphics[scale=0.45]{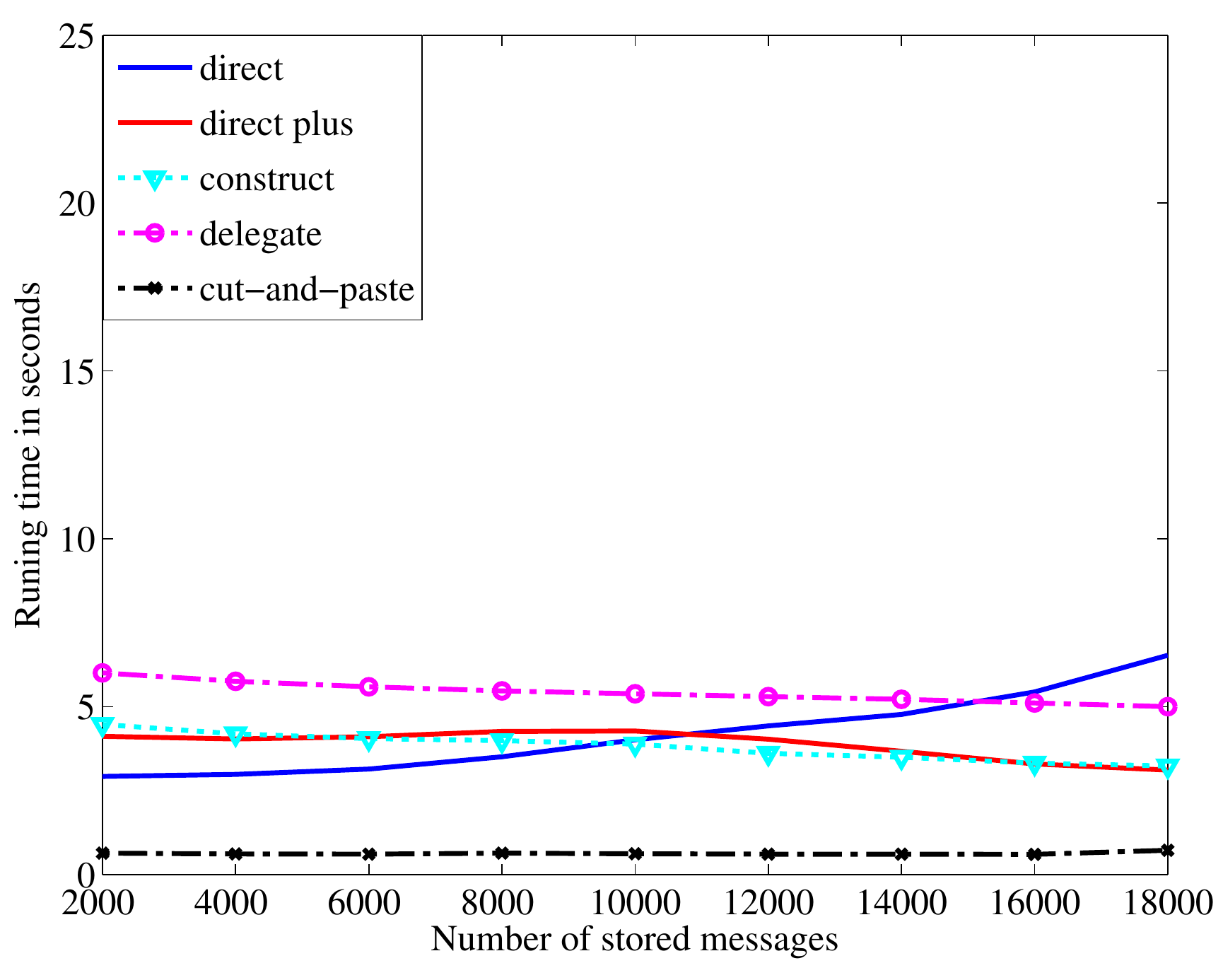}
\label{fig:fewlowtime}
}
\\
\subfloat[Many Low Probability Shift Errors]{
\includegraphics[scale=0.45]{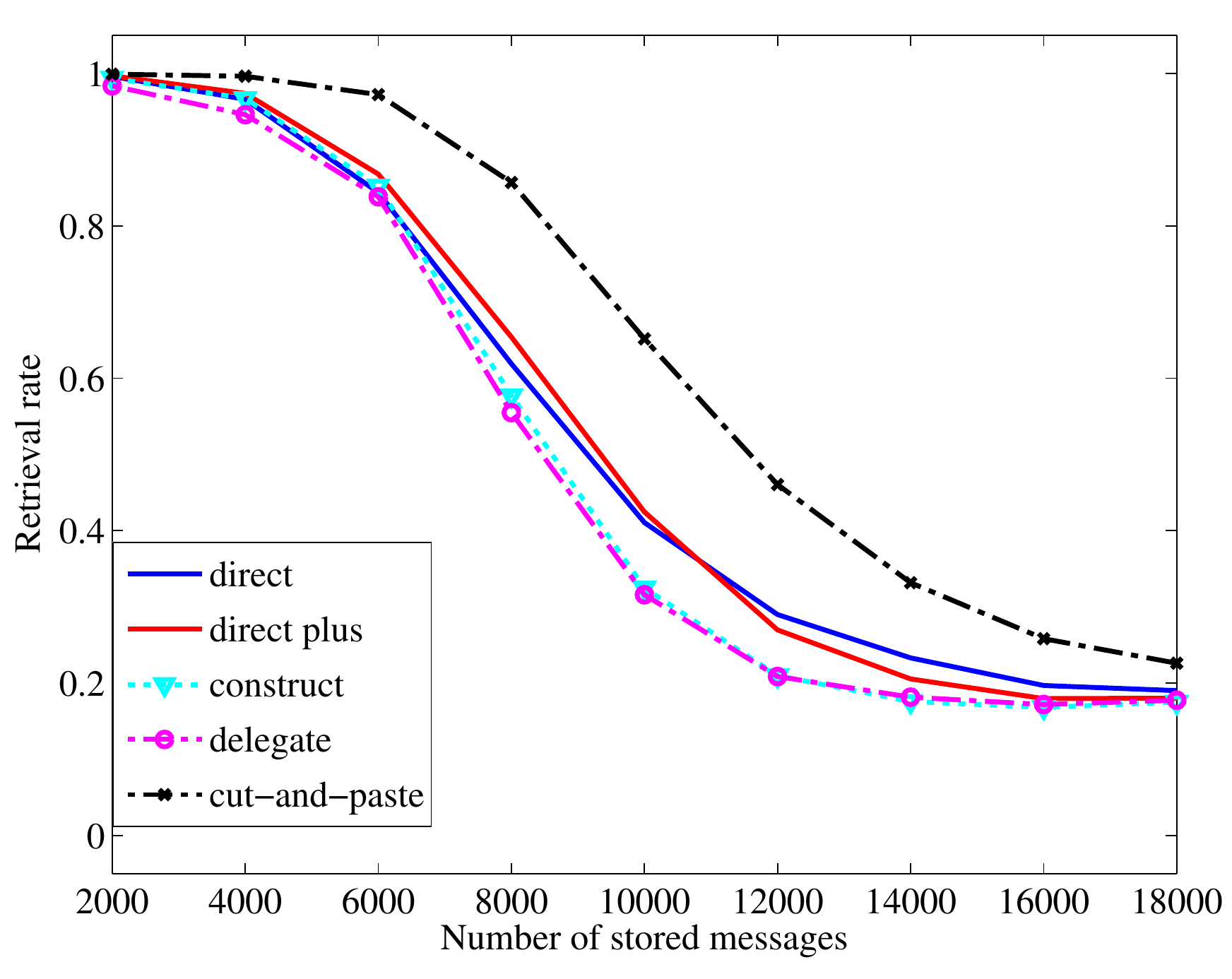}
\label{fig:manylowrate}
}
\subfloat[Many Low Probability Shift Errors]{
\includegraphics[scale=0.45]{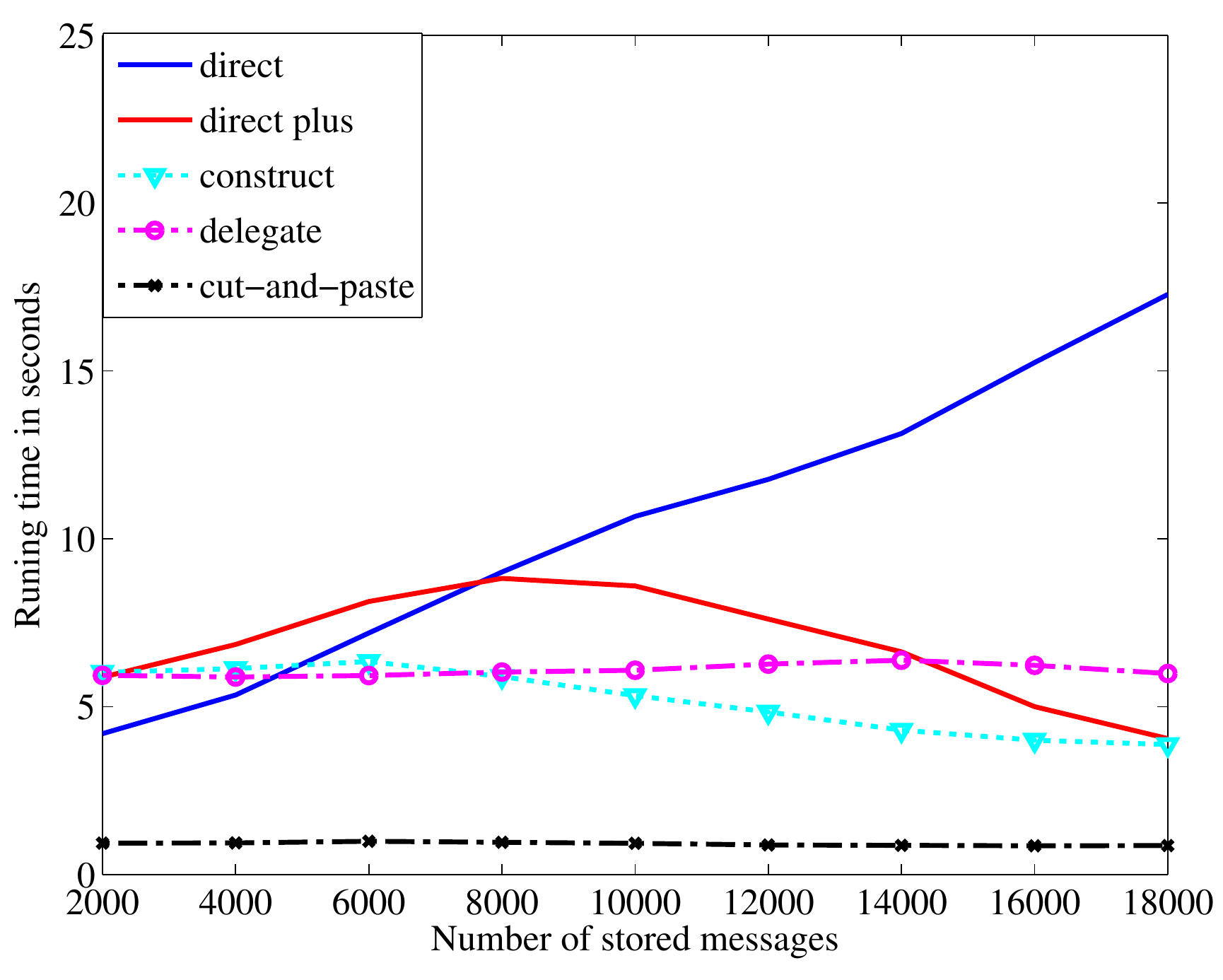}
\label{fig:manylowtime}
}
\\
\subfloat[Few High Probability Shift Errors]{
\includegraphics[scale=0.45]{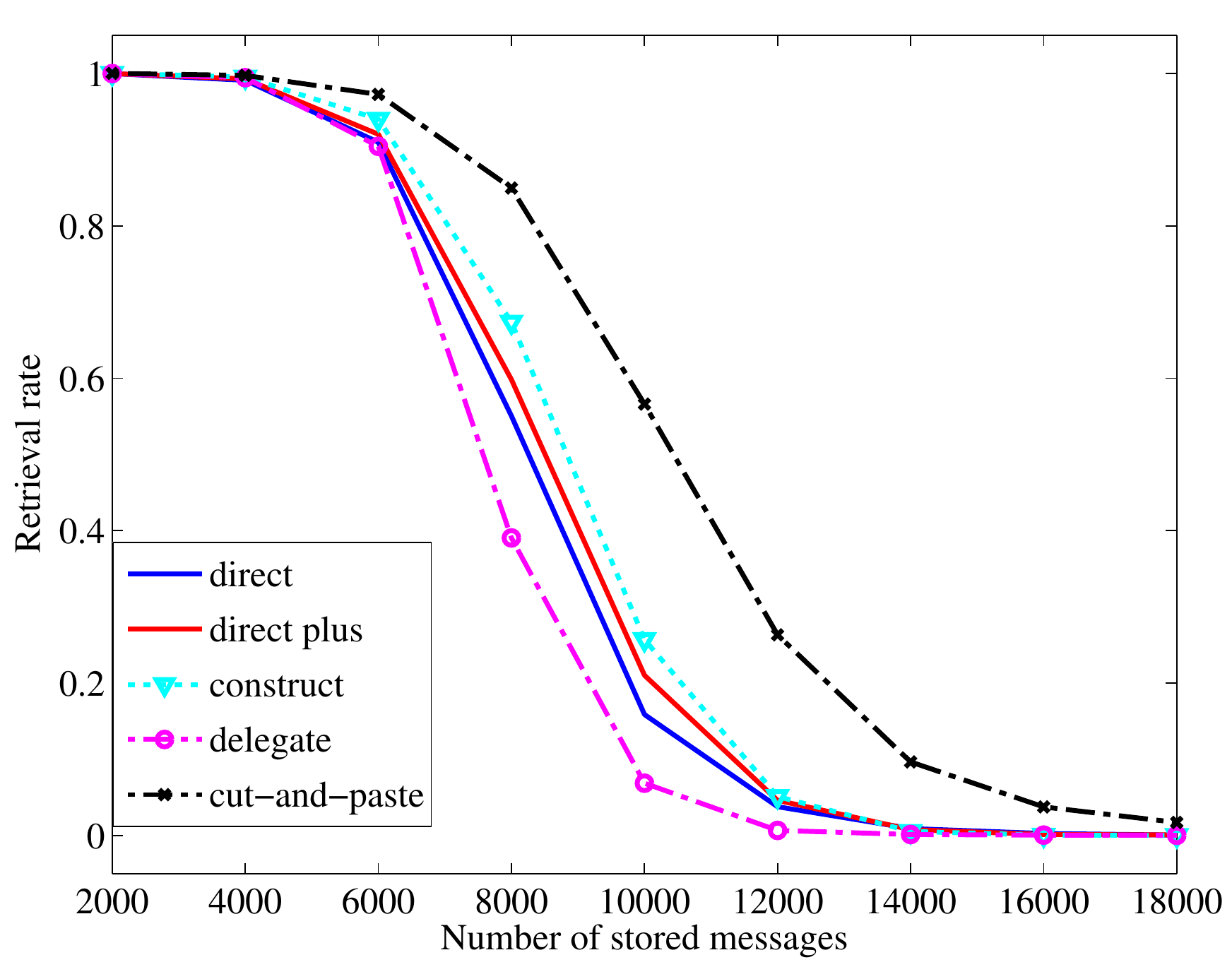}
\label{fig:fewhighrate}
}
\subfloat[Few High Probability Shift Errors]{
\includegraphics[scale=0.45]{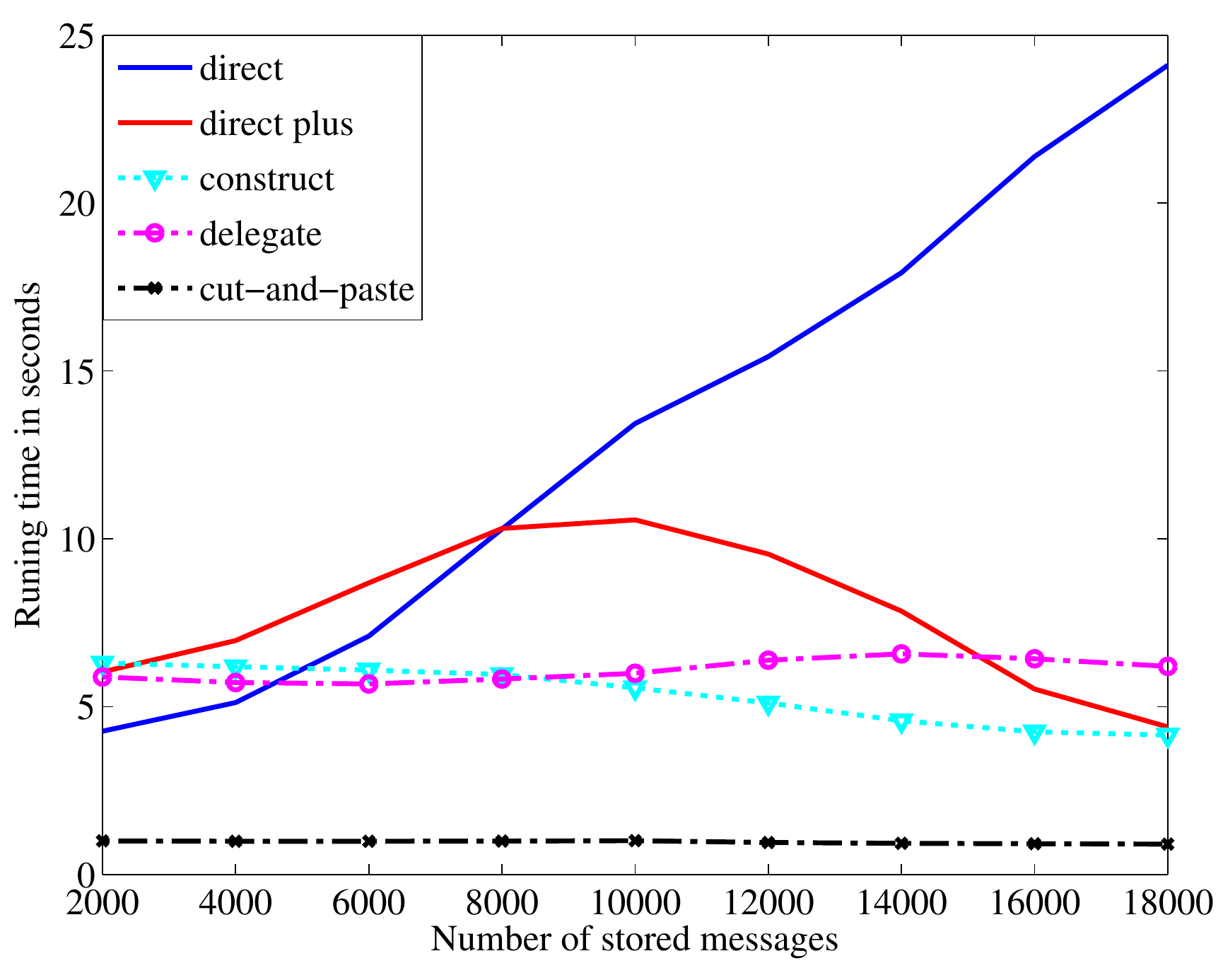}
\label{fig:fewhightime}
}
\caption{Comparisons of different approaches proposed in Section~\ref{sec:error} under shift errors. For \protect\subref{fig:fewlowrate} and \protect\subref{fig:fewlowtime}, two symbols are subjected to shift errors with probability $\frac{1}{5}$ independently. For \protect\subref{fig:manylowrate} and \protect\subref{fig:manylowtime}, all symbols are subjected to shift errors with probability $\frac{1}{5}$. For \protect\subref{fig:fewhighrate} and \protect\subref{fig:fewhighrate}, two symbols are subjected to shift errors for sure.}
\label{fig:shift}
\end{figure*}

We implement the shift errors by altering each symbol at a specified probability independently.
Shift errors may deteriorate in two ways as well: (1) The number of affected symbols is increasing. (2) The probability of altering symbols is increasing.
\fig{fig:shift} compares all approaches under these trends.
For \fig{fig:fewlowrate} and \fig{fig:fewlowtime}, two symbols are subjected to shift errors with probability $\frac{1}{5}$ independently.
For \fig{fig:manylowrate} and \fig{fig:manylowtime}, all symbols are subjected to shift errors with probability $\frac{1}{5}$.
For \fig{fig:fewhighrate} and \fig{fig:fewhightime}, two symbols are subjected to shift errors for sure.

We see from \fig{fig:shift} that the cut-and-paste approach again tops both in retrieval rate and running time.
The direct and direct plus approach perform more or less the same, with the direct plus approach winning the running time.
The delegate approach does not perform well under shift errors, even worse than the direct approach.

\subsubsection{All Errors Combined}
\begin{figure*}
\centering
\subfloat[]{
\includegraphics[scale=0.45]{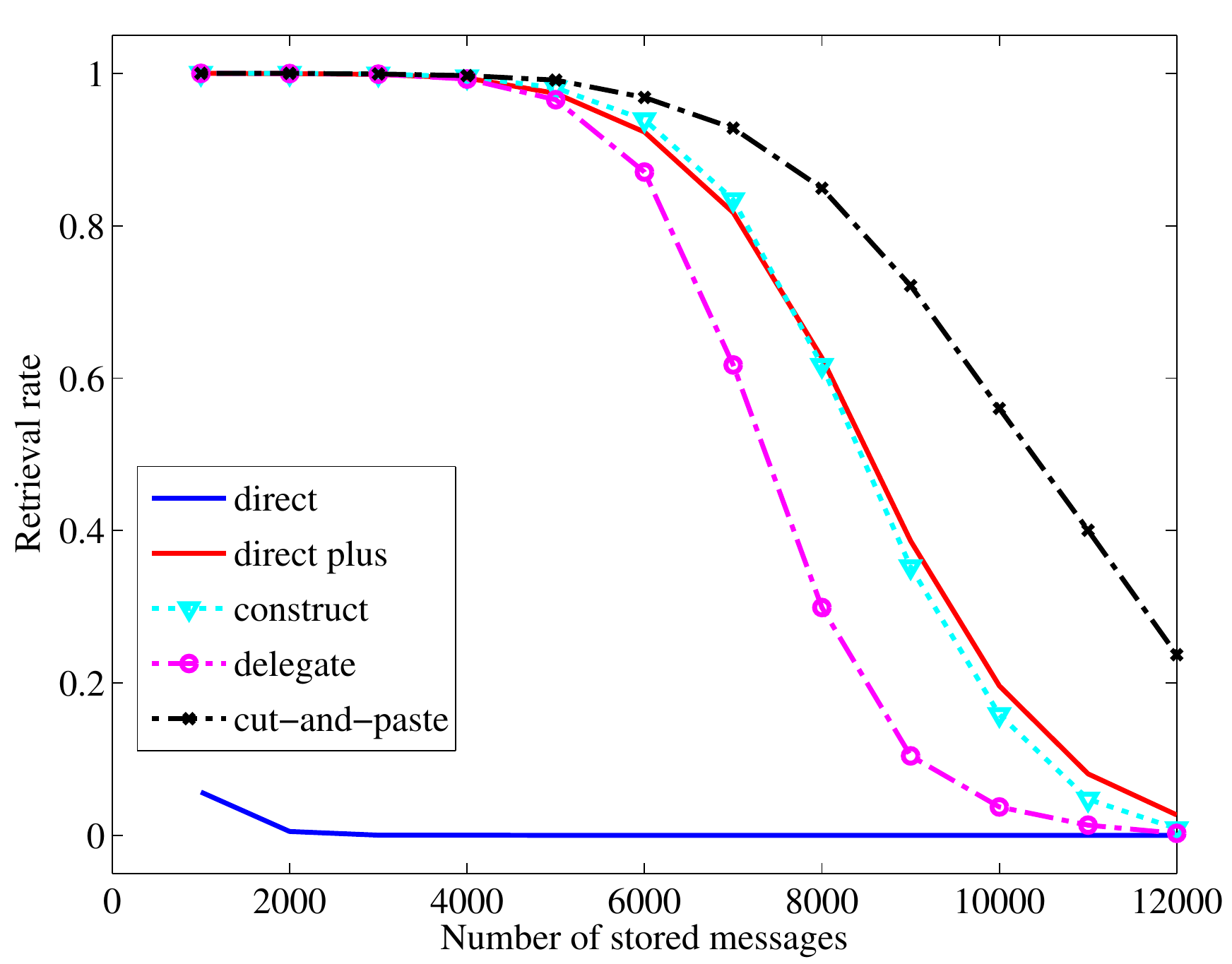}
\label{fig:rates}
}
\subfloat[]{
\includegraphics[scale=0.45]{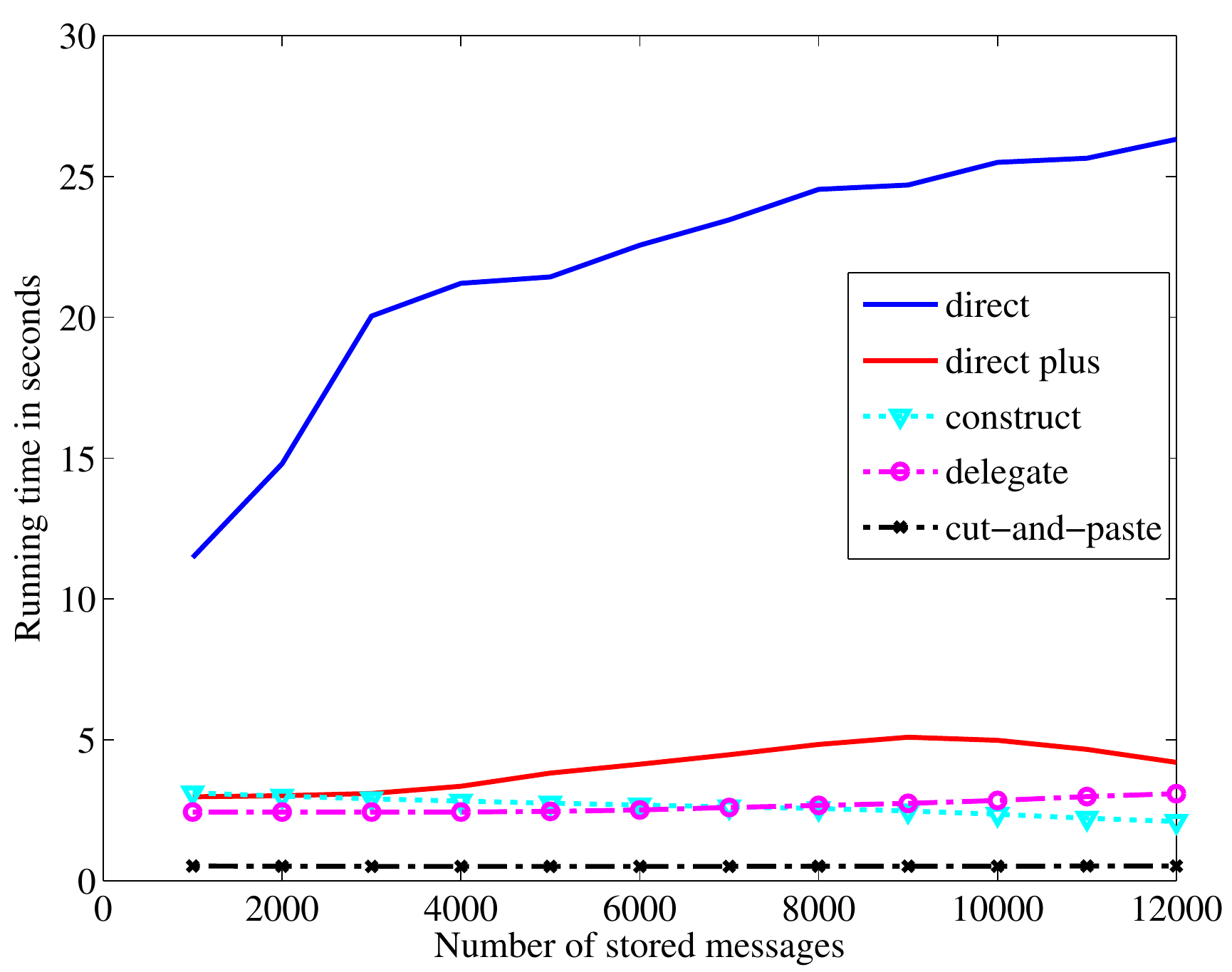}
\label{fig:times}
}
\caption{Comparison among all proposals when different errors are mixed together. The first three symbols are subjected to insertion errors. The last symbol experiences omission errors. The rest of the symbols have shift error with probability $\frac{1}{2}$.}
\label{fig:allerrors}
\end{figure*}

In this part, we compare the proposals when different errors are mixed together.
All symbols suffer from shift errors where each symbol is altered with probability $\frac{1}{2}$ independently.
The first three clusters are subjected to insertion errors with the first neuron in each cluster activated.
The last cluster experiences omission errors with every neuron deactivated.
Both the retrieval rate and running time are plotted in \fig{fig:allerrors}.

We see from \fig{fig:rates} that directly applying \sos{} performs the worst comparing with any other approaches, meanwhile the running time increases with the number of stored messages in \fig{fig:times}.
The construct and direct plus approach performs almost the same in terms of retrieval rate, which wins over the delegate approach.
The cut-and-paste approach performs excitingly well, e.g., when 12000 messages are stored in the network, it can still correctly retrieve more than 20\%, whereas the other approaches retrieve nothing.
To our surprise, although cut-and-paste is built purely on time demanding clique finding procedures, the running time is exceptionally fast in \fig{fig:times}.
We vary the probability of shift errors, the result is consistent with the comparison in \fig{fig:allerrors}.

We conclude that the cut-and-paste approach outperforms all the rest proposals in both retrieval rate and running time notably.
It should be considered as the default retrieving algorithm for CSAMs.

\subsection{USPS Dataset}
\begin{figure}
\centering
\includegraphics[scale=.5]{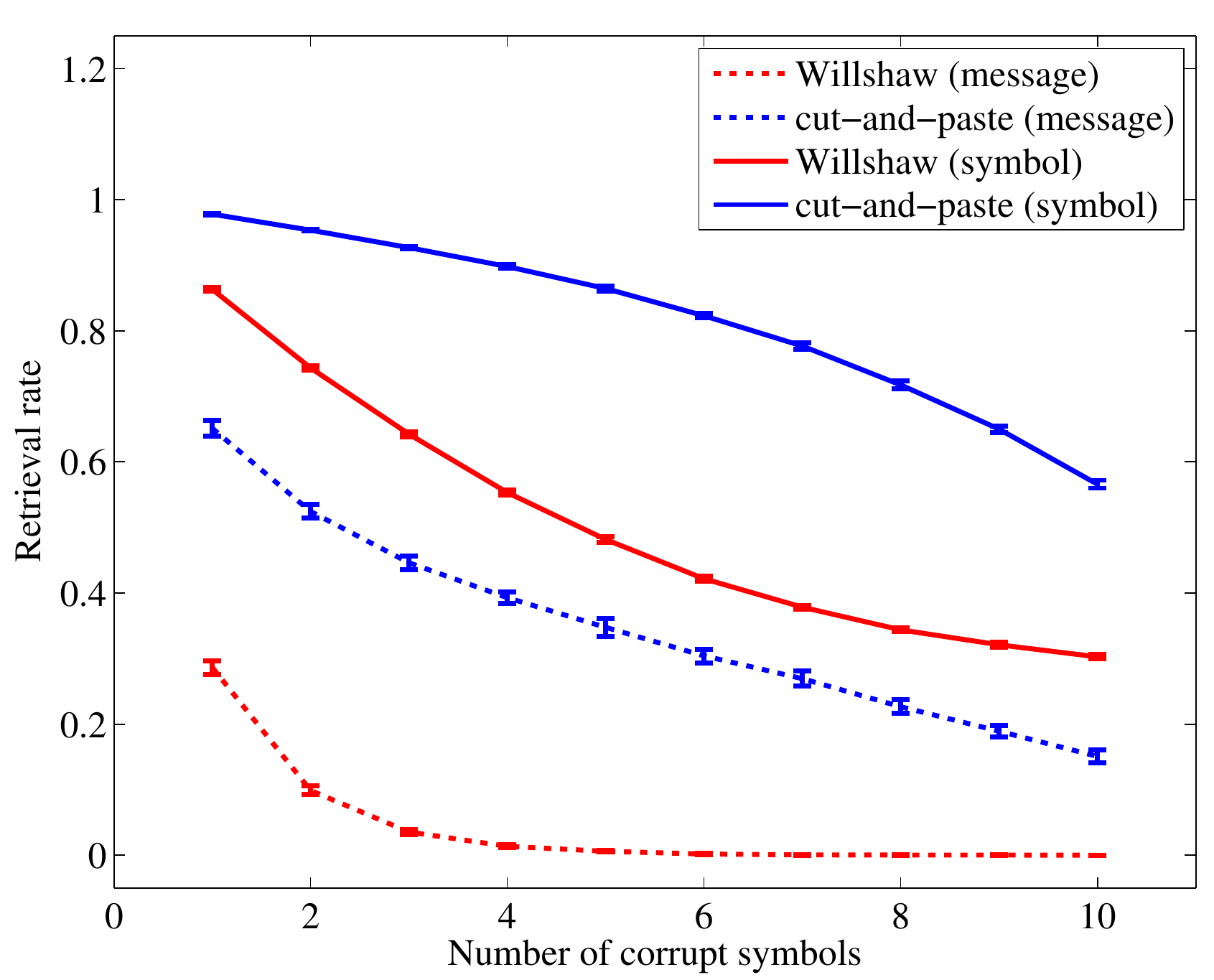}
\caption{The symbol retrieval rates and message retrieval rate comparison between the CSAM and the celebrated Willshaw network on the USPS dataset. Each $16\times 16$ USPS image is treated as two messages. We consider 8 successive pixels as a symbol. Therefore, $C=16$, $L=256$.}
\label{fig:usps}
\end{figure}

All the simulated data above are uniformly distributed, which is not a realistic assumption for real world applications.
In this sub-section, we test the effectiveness of CSAMs with the celebrated Willshaw network upon which many associative memory systems are constructed, using the famous USPS dataset.
In this dataset, 10 digits from 0 to 9 are taken from US postal envelopes, and each of them has 1100 image samples of the size $16\times 16$.
To make a fair comparison, we do not allow any fancy pre-processing to extract useful features except for binarizing the images to black and white.
We treat each image as 2 messages, and every 8 successive pixels as a symbol.
Therefore, $L=2^8=256$ and $C=\frac{16\times 16}{8*2}=16$.
This is a challenging task since the dataset itself is highly skewed, and also because one message is only regarded as successfully retrieved only when all 16 symbols are recovered.

Out of the 11000 images, we randomly sample 5000 as our stored messages ($10000$ messages in total), and then corrupt 1000 images as the probes into the network.
In \fig{fig:usps}, we run the experiment 10 times to average out the retrieval rate as a function of the number of corrupt clusters (symbols).
The standard deviation is also plotted for each data point.

\begin{figure*}
\centering
\subfloat[]{
\includegraphics[scale=1.5]{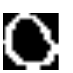}
}
\subfloat[]{
\includegraphics[scale=1.5]{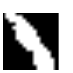}
}
\subfloat[]{
\includegraphics[scale=1.5]{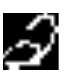}
}
\subfloat[]{
\includegraphics[scale=1.5]{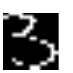}
}
\subfloat[]{
\includegraphics[scale=1.5]{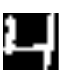}
}
\subfloat[]{
\includegraphics[scale=1.5]{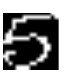}
}
\subfloat[]{
\includegraphics[scale=1.5]{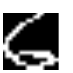}
}
\subfloat[]{
\includegraphics[scale=1.5]{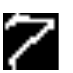}
}
\subfloat[]{
\includegraphics[scale=1.5]{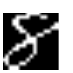}
}
\subfloat[]{
\includegraphics[scale=1.5]{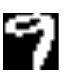}
}
\\
\subfloat[]{
\includegraphics[scale=1.5]{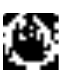}
}
\subfloat[]{
\includegraphics[scale=1.5]{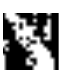}
}
\subfloat[]{
\includegraphics[scale=1.5]{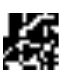}
}
\subfloat[]{
\includegraphics[scale=1.5]{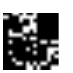}
}
\subfloat[]{
\includegraphics[scale=1.5]{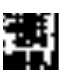}
}
\subfloat[]{
\includegraphics[scale=1.5]{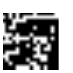}
}
\subfloat[]{
\includegraphics[scale=1.5]{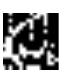}
}
\subfloat[]{
\includegraphics[scale=1.5]{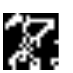}
}
\subfloat[]{
\includegraphics[scale=1.5]{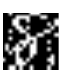}
}
\subfloat[]{
\includegraphics[scale=1.5]{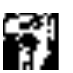}
}
\\
\subfloat[]{
\includegraphics[scale=1.5]{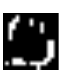}
}
\subfloat[]{
\includegraphics[scale=1.5]{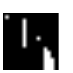}
}
\subfloat[]{
\includegraphics[scale=1.5]{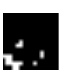}
}
\subfloat[]{
\includegraphics[scale=1.5]{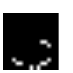}
}
\subfloat[]{
\includegraphics[scale=1.5]{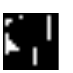}
}
\subfloat[]{
\includegraphics[scale=1.5]{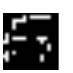}
}
\subfloat[]{
\includegraphics[scale=1.5]{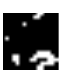}
}
\subfloat[]{
\includegraphics[scale=1.5]{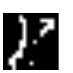}
}
\subfloat[]{
\includegraphics[scale=1.5]{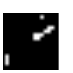}
}
\subfloat[]{
\includegraphics[scale=1.5]{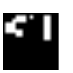}
}
\\
\subfloat[]{
\includegraphics[scale=1.5]{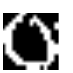}
}
\subfloat[]{
\includegraphics[scale=1.5]{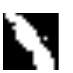}
}
\subfloat[]{
\includegraphics[scale=1.5]{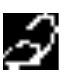}
}
\subfloat[]{
\includegraphics[scale=1.5]{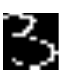}
}
\subfloat[]{
\includegraphics[scale=1.5]{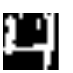}
}
\subfloat[]{
\includegraphics[scale=1.5]{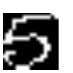}
}
\subfloat[]{
\includegraphics[scale=1.5]{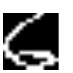}
}
\subfloat[]{
\includegraphics[scale=1.5]{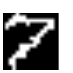}
}
\subfloat[]{
\includegraphics[scale=1.5]{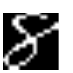}
}
\subfloat[]{
\includegraphics[scale=1.5]{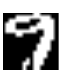}
}
\caption{
	Example plots of digit images with CSAMs and Willshaw networks.
	The first row are original stored images. The second row are corrupt probes with 5 symbols (40 pixels) damaged. The third row is for the Willshaw network. The last row is for the cut-and-paste rule of the CSAM.
}
\label{fig:digit}
\end{figure*}

Note that CSAMs outperform the Willshaw network considerably in terms of both the message retrieval rate and the symbol retrieval rate across different numbers of corrupt symbols.
For instance, when 4 out of 16 symbols are corrupt, the Willshaw network can hardly retrieve a complete image, whereas the CSAM is able to answer 40\% of the queries correctly.
The small deviations of the plots indicate the performance of the CSAM is stable and robust.
To better visualize the results, we also plot some example images in \fig{fig:digit}.
The first row contains the original images.
The second row is the corrupt probes, with 5 symbols (i.e., 40 pixels) being contaminated.
The third row is for the retrievals of the Willshaw network.
The last row is for the cut-and-paste approach of the CSAM.
It is not difficult to tell that the retrievals from CSAMs are superior than those from the Willshaw network.

\captionsetup[subfigure]{labelformat=empty}

\section{Summary\label{sec:summary}}
CSAMs are recently invented recurrent associative memories with a partite cluster structure featuring iterative retrieving schemes.
Two basic retrieving algorithms exist, namely \sos{} and \som{}.
Due to its close connection to a LDPC-like sparse binary coding, it should be resilient towards noise and errors in nature.
However, in former work, most attention is paid to the erasure scenario where a partial message is given as the probe to the network.
In this work, we extend CSAMs to cope with corruption scenarios where a corrupt probe is present.

We categorize and analyze the source of different errors when using an CSAM, so that the treatment to both erasure and corruption scenarios can be unified.
Then we make an amendment to the network structure.
Previously, we assume the incident degree as a measure of a neuron's popularity.
Unfortunately, this is problematic when the network becomes saturated.
At the cost of an extra scalar per neuron, we are able to fix this bias and improve further the retrieval rate.

Five different retrieving schemes are proposed to deal with corrupt probes.
We implement all algorithms and compare them under different types of errors separately to better understand the pros and cons of each scheme.
Out of all approaches, cut-and-paste performs notably better than the others in both the retrieval rate and running time simultaneously, except for the rare case when all neurons in a few clusters activate, which makes the running time significantly worse.
We suggest cut-and-paste be the default retrieving algorithm for CSAMs.
The USPS dataset is also tested to demonstrate the effectiveness of CSAMs, compared with the Willshaw network, where an exciting improvement is witnessed.

\section*{Acknowledgement}
\addcontentsline{toc}{section}{Acknowledgement}
This work was funded, in part, by the Natural Sciences and Engineering Research Council of Canada (NSERC), the \emph{Fonds Qu\'{e}b\'{e}cois de la recherche sur la nature et les technologies} (FQRNT) and the European Research Council project NEUCOD.
\bibliographystyle{IEEEtran}
\bibliography{my}

\newpage\onecolumn
\appendix[Two Building Modules for Algorithms Retrieving Corrupt Messages\label{sec:appendix}]
\begin{singlespace}
\begin{algorithm*}
\caption{\textbf{FindClique}, a modified version of Algorithm 2 in~\cite{yao2013bogus}, with the sorting measure at line 11 changed from node degree to the frequency vector $f$ amended in Section~\ref{sec:amendment}.}
\label{alg:kp}
\DontPrintSemicolon
\SetKwFunction{clique}{clique}
\SetKwFunction{update}{update}
\SetFuncSty{textbf}
\SetDataSty{textit}
\SetKwData{found}{found}
\SetKwData{level}{level}
\SetKwData{subgraph}{subgraph}

\KwIn{The frequency vector $f$ and the CSAM structure $R$ with $\card{R}$ clusters after reaching the bogus fixed point}
\KwOut{$Q$ (an active clique)}
\BlankLine\BlankLine\BlankLine
\textbf{global} $Q$ and \found\;
$Q \gets \emptyset$ and \found$\gets$ \textbf{false}\;
\clique{$R$}\;
\Return{$Q$}
\BlankLine\BlankLine\BlankLine
{\bf function} \clique{$U$}:\;
\Indp$\level\gets\card{Q}$\;
\If{$\level=\card{R}$}{
	$\found\gets\textbf{true}$\;
	\Return
}
sort $U_{\level}$ according to the frequencies of its nodes in the ascending order\footnotemark\;
\While{$U_{\level}\neq\emptyset$}{
	\If{any of $\lbrace U_{level+1}, \cdots, U_{\card{R}-1}\rbrace = \emptyset$}{
		\Return
	}
	$v\gets U_{level}[1]$\;
	$U_{level}\gets U_{level}\setminus\lbrace v\rbrace$\;
	$Q\gets Q\cup\lbrace v\rbrace$\;
	$\subgraph\gets\update{U, v, \level}$\;
	\clique{\subgraph}\;
	\If{\found}{
		\Return
	}
	$Q\gets Q\setminus\lbrace v\rbrace$\;
}
\BlankLine\BlankLine\BlankLine
\Indm{\bf function} \update{$U,v,\level$}:\;
\Indp\For{$i\gets level+1\;\textbf{to}\;\card{R}-1$}{
	$U_i\gets U_i\cap N(v)$\;
}
sort $\lbrace U_{level+1}, \cdots, U_{\card{R}-1}\rbrace$ according to the number of neurons in each set\;
\Return $U$
\end{algorithm*}

\footnotetext{At first glance, no matter according to node degrees in~\cite{yao2013bogus} or to frequencies in this work, it is suspicious to sort in the ascending order, which considers less appearing neurons first.
However, this setting does not only accelerate the recursive searching but also improve the retrieval rate.
It is easy to visualize that less appearing neurons prune the search tree in the early stage, which leads to significantly shorter running time.
For the second claim, although more appearing neurons associate with frequent messages, they also associate with more possible cliques.
The probability we retrieve the correct one decreases.
This error dominates, especially in saturated networks.
Therefore, the frequency information cannot be used by naively reverse the sorting order.}

\begin{algorithm*}
\caption{\textbf{Joint}, the joint scheme proposed in~\cite{yao2013gpugbnn}.}
\label{alg:joint}
\DontPrintSemicolon
\SetFuncSty{textbf}
\SetDataSty{textit}
\SetKwData{dead}{dead}
\SetKwData{signal}{signal}
\KwIn{The adjacency matrix $W$, the probe vector $v^0$ and the set of the missing clusters $P$.}
\KwOut{A fixed point $v$, possibly a bogus fixed point}
\BlankLine\BlankLine\BlankLine\BlankLine
\tcp{The joint scheme runs one iteration of \sos{} at the beginning.}
$s\gets W\cdot v^0$\;
\ForEach{$p \in P$}{
	\ForEach{$l$}{
		$v_{p,l}\gets 1\;\textbf{if}\;s_{p,l}=C-\card{P}\;\textbf{else}\;0$\;
	}
}
\BlankLine\BlankLine\BlankLine\BlankLine
\tcp{Afterwards, we run an optimized variant of \som{} until convergence.}
$v'\gets v^0$\;
\Repeat{$v = v'$}{
	$v\gets v'$\;
	\tcp{Only missing clusters are investigated.}
	\ForEach{$p\in P$}{
	\ForEach{$l'$}{
		$k=(p-1)L+l'$\tcp*[f]{Convert to global index.}\;
		\eIf{$v_k=0$}{$v'_k\gets 0$\tcp*[f]{Only active neurons are investigated; see~\cite{yao2013gpugbnn}.}}{
			$\dead\gets\textbf{false}$\;
			\ForEach{$c$}{
				$\signal\gets\textbf{false}$\;
				\ForEach{$l$}{
					\If	{$W_{k,(c-1)L+l}=1\;{\bf and}\;v_{(c-1)L+l}=1$}{
						$\signal\gets\textbf{true}$\;
						$\textbf{break}$\;
					}
				}
				\If{$\signal={\bf false}$}{
					$\dead\gets\textbf{true}$\;
					$\textbf{break}$\;
				}
			}
			$v'_k\gets 0\;\textbf{if}\;\dead\;\textbf{else}\;1$\;	
		}
	}
	}
}
\end{algorithm*}
\end{singlespace}

\end{document}